\documentclass[3p]{elsarticle}
\pdfoutput=1
\usepackage[utf8]{inputenc}
\usepackage{graphicx}
\usepackage{subcaption}  % Keep only this, remove subfig
\usepackage{hyperref, amsmath, amsfonts, natbib, bm, floatrow, multirow, wrapfig}
\usepackage[dvipsnames]{xcolor}
\usepackage{ulem} 
\usepackage{algorithm}
\usepackage{algpseudocode}
\usepackage{caption}
\usepackage{subcaption}
\usepackage{float}
\usepackage{booktabs}
\usepackage{appendix}

\graphicspath{{Figures/}}

\journal{Journal name}

\begin{document}

\begin{frontmatter}
\begin{center}
  { \fontsize{16}{10} \textbf{Efficient Adaptation of ROMs for Unsteady Flows Using Data Assimilation} } \\
    \vspace{0.5cm}
    \normalsize \textbf{Ismaël Zighed$^{1,2, 3}$}, \textbf{Andrea Nóvoa$^{4,5}$, \textbf{Luca Magri}$^{4}$, \textbf{Taraneh Sayadi}$^{6}$} \\
  { \fontsize{9}{10}\selectfont  \textit{
  $^1$Institut Jean le Rond d'Alembert, Sorbonne Université, 
  $^2$ISIR, Sorbonne Université, $^3$ SCAI, Sorbonne Université, 
  $^4$Aeronautics department, Imperial College London, 
  $^5$I-X, Imperial College London,
  $^6$M2N, Conservatoire National des Arts et Metiers} } \\ % Example of superscript for multiple affiliations
  { \fontsize{9}{10}\textit{Corresponding author: Ismaeël Zighed, (ismael.zighed@sorbonne-universite.fr)}} \\
\end{center}

\begin{abstract}
    We propose an efficient fine-tuning strategy for a parametrized Reduced Order Model (ROM) that attains accuracy comparable to full retraining while requiring only a fraction of the computational time (from 2 hours to $\approx 15$ minutes) and relying solely on sparse observations of the full system of order $10^1$ to $10^2$. The architecture employs an encode–process–decode structure: a Variational Autoencoder (VAE) to perform dimensionality reduction, and a transformer network to evolve the latent states and model the dynamics. The ROM is parametrized by an external control variable, the Reynolds number in the Navier–Stokes setting, with the transformer exploiting attention mechanisms to capture both temporal dependencies and parameter effects. The probabilistic VAE enables stochastic sampling of trajectory ensembles, providing predictive means and uncertainty quantification through the first two moments. After initial training on a limited set of dynamical regimes, the model is adapted to out-of-sample parameter regions using only sparse data. Its probabilistic formulation naturally supports ensemble generation, which we employ within an ensemble Kalman filtering framework to assimilate data and reconstruct full-state trajectories from minimal observations. We further show that, for the dynamical system considered, the dominant source of error in out-of-sample forecasts stems from distortions of the latent manifold rather than changes in the latent dynamics. Consequently, fine-tuning can be limited to the autoencoder, allowing for a lightweight, computationally efficient, near-real-time adaptation procedure with very sparse data. 
\end{abstract}

\end{frontmatter}
\section{Introduction}
\label{sec:intro}

Modelling physical systems using data-driven approaches remains a notoriously difficult task. Even for seemingly simple systems, the challenge arises due to (i) their nonlinear behaviour, high dimensionality, and complex space–time dependencies, and (ii) the fact that the available data are often sparse, noisy, and expensive to generate \citep{Kelshaw_Magri}. 

To address the first challenge, Reduced-Order Models (ROMs) are a powerful framework capable of transforming limited data into meaningful predictions by exploiting latent low-dimensional structures within the system, essentially mining simplicity from complexity. These low-dimensional structures, commonly referred to as manifolds, can be defined in several ways. They may exist in a space spanned by physical observables, as in a phase space~\citep{SSM-1,SSM-2}, or in a basis of modes derived from a linear projection onto a low-dimensional latent space that maximizes energy~\citep{POD_Schmid}. Alternatively, they can be identified in a purely data-driven manner by minimizing reconstruction error, as in nonlinear autoencoders, resulting in a highly compressed and abstract latent space. Remarkably, such low-dimensional manifolds can exist in chaotic systems~\citep{ChaoticSSM}. Once identified, the system dynamics can be projected and simplified. Methods to achieve this can be either intrusive~\citep{CD-ROM, ql-ROM} or non-intrusive~\citep{ILED, Gupta, VpROM}, i.e., data-driven.

Statistical mechanics have long emphasized the importance of memory, which plays a central role in recovering partial system states. This insight has repelled dynamical models from the attractor of Markovian determinism, which even finite Koopman subspaces often fail to represent accurately. The Mori-Zwanzig formalism and Takens’ theorem \citep{Taken,Mori,Zwanzig} emphasize the importance of incorporating a memory term to effectively account for the influence of the unobserved, orthogonal subspace on the set of observables learned or selected during the reduction. Additionally, studies aiming at uncovering finite Koopman spaces~\citep{Gupta} demonstrate the limitations of using a purely Markovian representation for rolling out dynamics within a finite, invariant Koopman space. The concept now has further influenced data-driven modelling. Notable examples include reservoir computing \cite{Real_time_calibration_chaotic, racca_2023}, embedded memory \cite{Gupta} and, most recently, attention mechanisms and transformers~\citep{Attention}. These developments have substantially reshaped the landscape of nonlinear dynamical modelling~\citep{SLT, AutoencoderTransformer,UPdROM}. 

Identifying suitable latent manifolds, however, remains nontrivial. An ill-defined manifold leads to error propagation from the latent dynamics to the system’s physical states, causing predictions to diverge from the true trajectories even after moderately long rollouts \cite{thermalizer}. Correcting this divergence, either in real time through control or a posteriori through fine-tuning, may therefore be necessary. However, both strategies typically require additional data, giving rise to a data bottleneck when incorporating a posteriori knowledge of the true system state into the model’s predictive process.

Most existing approaches assume access to high-fidelity, fully sampled simulation or experimental data for training and fine-tuning, an assumption that rarely holds in practice. During fine-tuning, we instead assume that the model’s prediction of the high-dimensional state is only partially correct. This partial correctness can be exploited in conjunction with sparse high-fidelity measurements, for instance, those obtained from physical sensors, to enhance model accuracy and robustness. The problem can also be viewed through a statistical lens, where the new, limited fine-tuning data are combined with imperfect prior estimates whose error covariances are explicitly accounted for. In this framework, inaccurate model predictions are not discarded but treated as uncertain observations, allowing their associated uncertainty to guide the integration of sparse, high-fidelity data. This principle underpins Kalman filtering~\cite{Kalman_original}, where an error-prone model estimate is combined with limited yet accurate observations and has been applied extensively in the data-assimilation community \cite{DA, bocquet_2}. When this condition holds, Kalman filters can be applied effectively across a broad range of contexts, including online modelling~\citep{online_model_learning,Real_time_calibration_chaotic}, model bias estimation~\citep{bias_aware_DA}, reinforcement learning~\citep{Control_turbulence_DA}, and a posteriori model retraining~\citep{bocquet_2}. 
Kalman filters are purely statistical in nature and do not depend on the physical properties of the underlying system. Consequently, they have been successfully employed in complex parametric spaces~\citep{parameter_augmented_DA} and in turbulent flow applications~\citep{turbulence_DA}. 

Our approach is inspired by~\citep{bocquet_2} in that the model is retrained a posteriori using fine-tuning data generated through Kalman-filter–based data assimilation. The dynamical model used in this work is a stochastic Reduced-Order Model (ROM), described in Section~\ref{subsec:ROM}, and builds upon previous developments that did not incorporate sparse-data constraints~\citep{UPdROM}. The stochastic nature of the ROM provides ensembles of predictions, whose variance is used as a proxy for model error. A central focus of this work is to alleviate the cost associated with generating fine-tuning data. To this end, the methodology leverages ensemble predictions within a Gaussian-based uncertainty-quantification framework. This inherent Gaussianity enables efficient reconstruction of full-state information from sparse, high-fidelity measurements distributed across the physical domain via Kalman filtering. The model, originally trained in a specific region of the testing distribution, is subsequently fine-tuned in a different region using only a minimal amount of additional data.
What distinguishes our method is the tight integration of probabilistic ROMs with Kalman filtering, specifically tailored for parametric dynamical systems. This integration allows the model to adapt across different parameter regimes, each characterized by distinct system behaviours, whereas traditional approaches would require costly, full-state, high-fidelity data to achieve comparable adaptability. \\

The paper is structured as follows. Section~\ref{sec:problem} introduces the experimental setup and \ref{subsec:ROM} presents the ROM used in this study. Section~\ref{sec:manifold} then investigates the training of the ROM, together with an analysis of the resulting latent space and manifold, and their impact on learning performance. This analysis reveals a remarkable economy in training requirements, as only the projection onto the latent space requires fine-tuning. Finally, this economy is exploited in Section~\ref{sec:DA}, where sparse observations and Kalman filtering are combined to efficiently retrain the model.
Conclusions and perspective of the work are presented in Section \ref{sec:conclu}. 

\section{Parametrised dynamical system}
\label{sec:problem}

\begin{figure}[!htb]
\hspace{-5em}
    \includegraphics[width=0.9\linewidth]{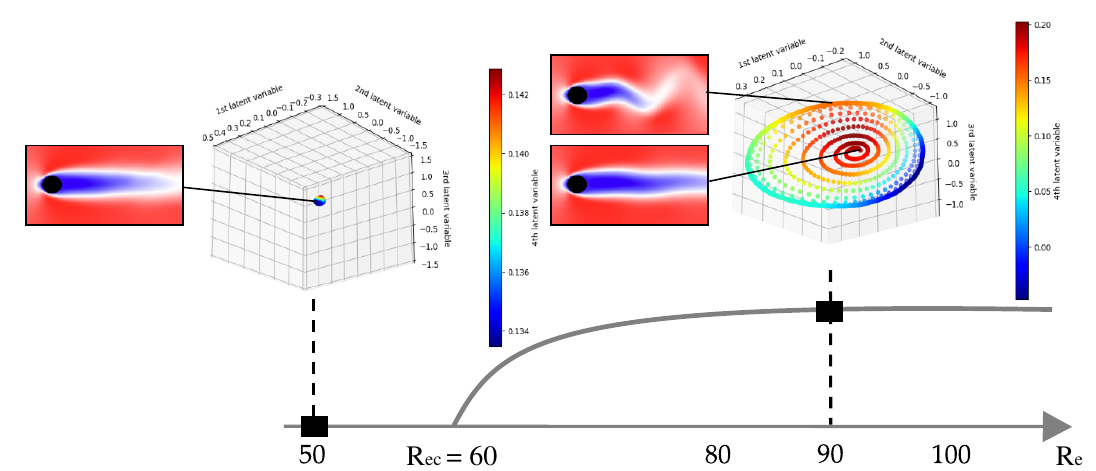}
    \caption{Hopf bifurcations with latent manifolds at $Re=50$ and $Re = 90$. $U$ velocity field at unstable fixed point and at limit cycle.}
    \label{fig:hopf}
\end{figure}

We consider the following equation, which describes a general form of a dynamical system
\begin{equation}
    \frac{d\psi}{dt} = F(\psi, \xi, t),
\end{equation}
where \( F \) is a nonlinear operator governing the spatial and temporal evolution of the system, \(\psi\) denotes the state variables, and \(\xi\) represents the parameters influencing the system's dynamical behaviour. We study the dynamical system of an unsteady, two-dimensional flow past an obstacle. Given this representation, the above equation can be recast as the incompressible Navier–Stokes equations, 
\begin{equation}
\label{eq:NS}
\begin{aligned}
\frac{\partial \mathbf{\psi}}{\partial t}
    &= -\nabla p - (\mathbf{\psi} \cdot \nabla) \mathbf{\psi} + \frac{1}{Re} \nabla^2 \mathbf{\psi}, \\
\nabla \cdot \mathbf{\psi}
    &= 0.
\end{aligned}
\end{equation}
Here, $\psi$ is the velocity field $[u,v]^T$ and $p$ the pressure field.
The flow behaviour is primarily influenced by the Reynolds number \( Re = \frac{\rho U L}{\mu}\), where \( \rho \) is the fluid density, \( U \) is the characteristic velocity, \( L \) is the characteristic length (such as the length of the bluff body), and \( \mu \) is the dynamic viscosity of the fluid. This quantity is treated here as the variable model parameter \( \xi \). 
Increasing the Reynolds number can cause the system to bifurcate, 
which leads to the onset of instabilities or transitions between periodic orbits. 
Prior studies on circular cylinders have characterized a primary bifurcation at $Re\sim 48$ followed by an instability growth rate linearly dependant to the Reynolds number~\citep{Canuto_Taira_2015, Strykowski_Sreenivasan_1990}. 
Here, this \textit{Hopf} bifurcation occurs at $Re \sim 60$ because the obstacle is an ellipsoid with an aspect ratio greater than one. This configuration offers the advantage of exhibiting a dynamical behaviour characterized by a single periodic orbit over a broader range of Reynolds numbers, as illustrated in Figure \ref{fig:hopf}. 

We focus on the moderate Reynolds number regime \( Re \in [80, 140] \), where the flow is characterized by a single periodic orbit. Although, to the best of our knowledge, no rigorous stability analysis has been reported for this specific configuration, our numerical observations strongly indicate that the Hopf bifurcation occurs at $Re=60$. The system is initialized at the unstable fixed point, which exhibits nearly stationary behaviour before transitioning to its stable limit cycle, i.e., the well-known Von Kármán vortex street forming in the wake of the obstacle (c.f., Fig.~\ref{fig:hopf}). 
%
% The model may struggle to generalize and extrapolate effectively across this range, particularly near the upper boundary, thus necessitating fine-tuning. Thus, the flow regime considered here lies beyond the onset of instability and is characterized by the emergence of a single periodic orbit. 
%\begin{figure}[!htb]
%    \centering
%    \includegraphics[width=0.7\linewidth]{Images/Part1/system_energy.pdf}
%    \caption{Initial and limit-cycle states of the system, along with the evolution of its kinetic energy.}
%    \label{fig:system_energy}
%\end{figure}
The system state is discretized on an Eulerian grid of size \( 131 \times 100 \), resulting in a high-dimensional state space of 13100 variables for the two velocity components \( u \) and \( v \). We use an Immersed Boundary Method solver (ibmos) \citep{IBMOS} to generate training data.

\section{Parametric Reduced Order Model (ROM)}
\label{subsec:ROM} 

Dynamical systems often converge towards low-dimensional structures, known as manifolds, on which the system's dynamics evolve. 
In this work, we take a parameter-dependent Reduced Order Model (ROM) approach that compresses the spatial information while simultaneously modelling the temporal dependencies within this reduced-dimensional manifold \citep{UPdROM}.
Specifically, we employ a parameterized Variational Autoencoder (VAE) \citep{VAE} to identify a reduced latent space onto which the dynamics are learned by a Transformer, leveraging recent advances in embedded memory and attention mechanisms \citep{Attention,SLT}. 
\begin{figure}[!htb]
\centering
\includegraphics[width=1\linewidth]{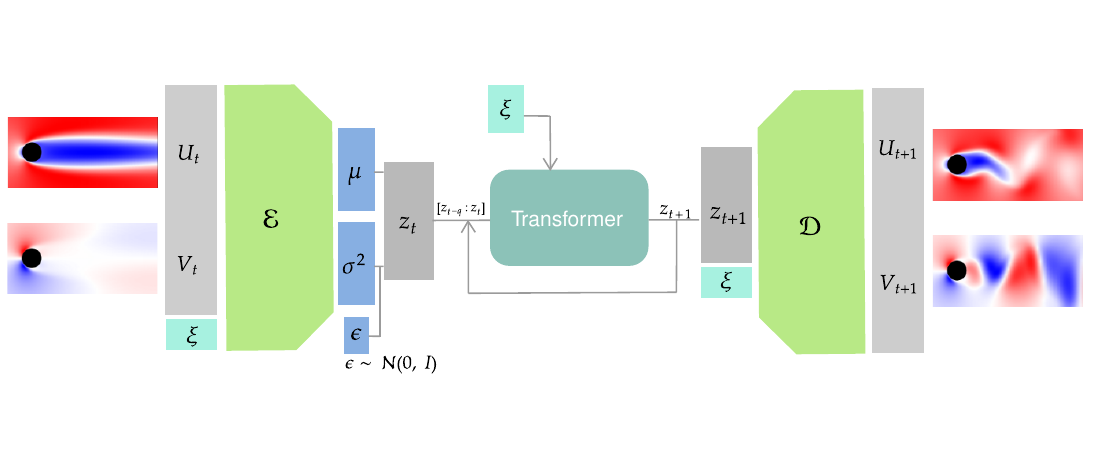}
\vspace{-4em}
\caption{Architecture of the parametric Reduced-Order Model encodes and reduces with $\mathcal{E}$ both velocity fields $U$ and $V$ along with the parametrisation $\xi$. The transformer evolves in time the dynamics auto-regressively in the latent space with self-attention and cross-attention with $\xi$. Finally the decoder $\mathcal{D}$ projects the latent states and $\xi$ back to the physical space.}
\label{fig:ROM}
\end{figure}

The proposed architecture, illustrated in Figure~\ref{fig:ROM}, is based on the  Uncertainty-Aware and Parametrised dynamic Reduced-Order Model (UP-dROM) \citep{UPdROM, GitHubRepo}. 
The model learns unsteady-state behaviour and performs dynamical time forecasting from initial conditions in an auto-regressive fashion. 
This architecture is built upon three key features: 
\begin{enumerate}
    \item \textit{Probabilistic framework}: The Variational Autoencoder (VAE) yields a well-structured and dense  latent space, which enhances model generalization, while learning a  distribution over latent variables to enable the generation of diverse, physically plausible trajectories. The VAE-based approach naturally supports stochastic sampling and uncertainty quantification \citep{UPdROM}. This probabilistic framework can be leveraged for ensemble Data Assimilation, as detailed in Sec. \ref{sec:DA}. 
    \item \textit{Parameterization}: To account for the fact that different flow regimes may occupy distinct regions of the latent manifold \citep{ql-ROM}, all the components of the ROM, namely Encoder, Transformer, and Decoder, are explicitly conditioned on the Reynolds number $Re$. 
    \item \textit{Latent attention}: The Transformer evolves the system dynamics within the latent space, utilizing self-attention to capture long-range temporal dependencies in the vortex street and cross-attention to adapt the evolution to specific $Re$. This latent-space approach ensures temporal order through positional encoding while providing the scalability required to resolve the complex dynamics of high-dimensional state spaces. 
\end{enumerate}

For training, the original $131 \times 100$ grid is flattened to a $13{,}100$-dimensional vector and downsampled by a factor of 4, yielding $3{,}275$ degrees of freedom. Since both velocity components $(u,v)$ are modeled simultaneously, the resulting input space lies in $\mathbb{R}^{6{,}550}$.  
The VAE then learns a mapping $\mathcal{E}$
\[
\mathcal{E} : \mathbb{R}^{6{,}550} \rightarrow \mathbb{R}^{4},
\]
with the Transformer trained jointly on the latent representations. Additional details on the architecture and hyperparameters are provided in ~\ref{subsec:ROMarch}.

\subsection{Model performance across the parametric region}
First, we train the model only at $Re = 90$ and $Re = 120$ and test the model over $Re \in 
\{80, 90, \dots, 130, 140\}$, thereby covering both interpolation and extrapolation regimes. The dataset consists of $T = 3033$ snapshots, with an input dimension of $m = 6550$. We split the dataset into training and testing sets by assigning even snapshots to the training set and odd snapshots to the test set to ensure equal share of transients and limit cycle to both datasets. Accordingly, the training set has dimensions $\mathbb{R}^{2, T/2, m}$, and the test set also has dimensions $\mathbb{R}^{2, T/2, m}$.  Different systems might benefit from a contiguous temporal split rather than the odd-even strategy adopted here. However, for the present system, such a split would result in the model being trained primarily on transient dynamics while being evaluated mostly on the simple periodic limit-cycle regime. This would lead to a biased evaluation procedure.
Finally, the validation set has dimensions $\mathbb{R}^{7, T/2, m}$, containing system states for all considered Reynolds numbers. The validation set is designed to explore the parameter space, enabling an assessment of the model’s generalization performance, even though the model was trained only in a sub-region of this space.  

Kinetic energy is used to evaluate the performance of the model. It aggregates both velocity components $(u, v)$ into a single physically meaningful quantity, and for a given parameter $\xi$ is computed as
\begin{equation}\label{eq:K}
    k_\xi = u_\xi^2 + v_\xi^2, \quad 
    k_\xi \in \mathbb{R}^{T/2, D}.
\end{equation}
In addition, model performance is assessed in forecast mode on the test set using two metrics: (i) the  energy distance, and (ii) uncertainty quantification (UQ). 
\begin{figure}[!htb]
    \centering
    \includegraphics[width=1\linewidth]{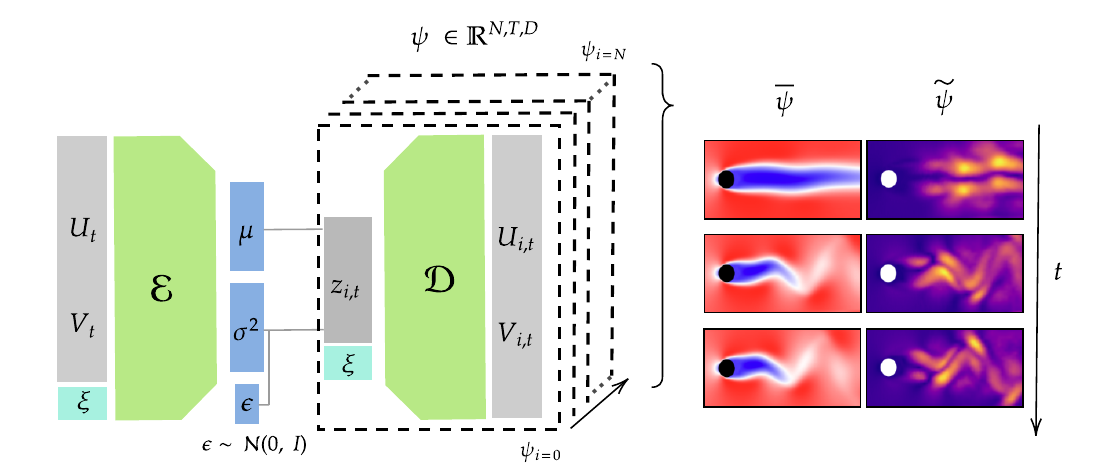}
    \caption{ROM uncertainty evaluation through ensemble generation $\psi$. The ensemble mean (prediction) is denoted $\bar{\psi}$ and the ensemble variance (uncertainty) by $\tilde{\psi}$.}
    \label{fig:UQ}
\end{figure}
%s
%
First, the energy distance metric treats the predicted and true trajectories as probability distributions rather than pointwise deterministic signals. For two random variables $X \sim P$ and $Y \sim Q$ on $\mathbb{R}$, the energy distance between the corresponding probability measures $P$ and $Q$ is defined as : 
\[
D_E(P,Q)
=
\left(
2\,\mathbb{E}\|X-Y\|
-
\mathbb{E}\|X-X'\|
-
\mathbb{E}\|Y-Y'\|
\right)^{1/2},
\]
where $X'$ and $Y'$ are independent copies of $X$ and $Y$, respectively. In the one-dimensional case, the energy distance is equivalently related to the integrated squared difference between cumulative distribution functions,
\[
D_E^2(P,Q)
=
2 \int_{-\infty}^{+\infty}
\left(
F_P(t)-F_Q(t)
\right)^2 dt,
\]

Since our signal is defined over $m$ spatial dimensions, the energy distance is computed independently for each dimension and then averaged to obtain a global discrepancy measure.

Unlike pointwise Euclidean metrics, the energy distance compares the statistical distributions of the trajectories rather than enforcing an exact temporal alignment. Consequently, it remains robust to small phase shifts between the predicted and reference signals. This property is particularly relevant for oscillatory dynamical systems, where the reduced-order model may accurately reproduce the amplitude and frequency content of the dynamics while exhibiting a slight temporal lag or lead. In such situations, classical $\ell_2$-based errors can severely penalize otherwise dynamically consistent predictions, whereas the energy distance provides a smoother and more statistically meaningful assessment of trajectory similarity~\citep{Wasserstein-phaseshift-robustness}. Furthermore, because the metric relies on distributional discrepancies, it is generally less sensitive to localized noise fluctuations and better reflects global statistical agreement between trajectories.
Second, Uncertainty quantification (UQ) is computed as the variance of the stochastically sampled ensemble drawn from the latent coordinate distributions, as illustrated in Figure~\ref{fig:UQ} for an ensemble of size $N$. The variance is then averaged over the entire forecast window to obtain a scalar value associated with each Reynolds number configuration. The general UQ evaluation framework is described in detail in \citep{UPdROM}. Importantly, the model error and its estimated uncertainty are expected to be correlated. This correlation serves as a crucial sanity check for both our uncertainty quantification strategy and our error evaluation metric. 
\\

\begin{figure}[!htb]
    \centering
    \includegraphics[width=0.9\linewidth]{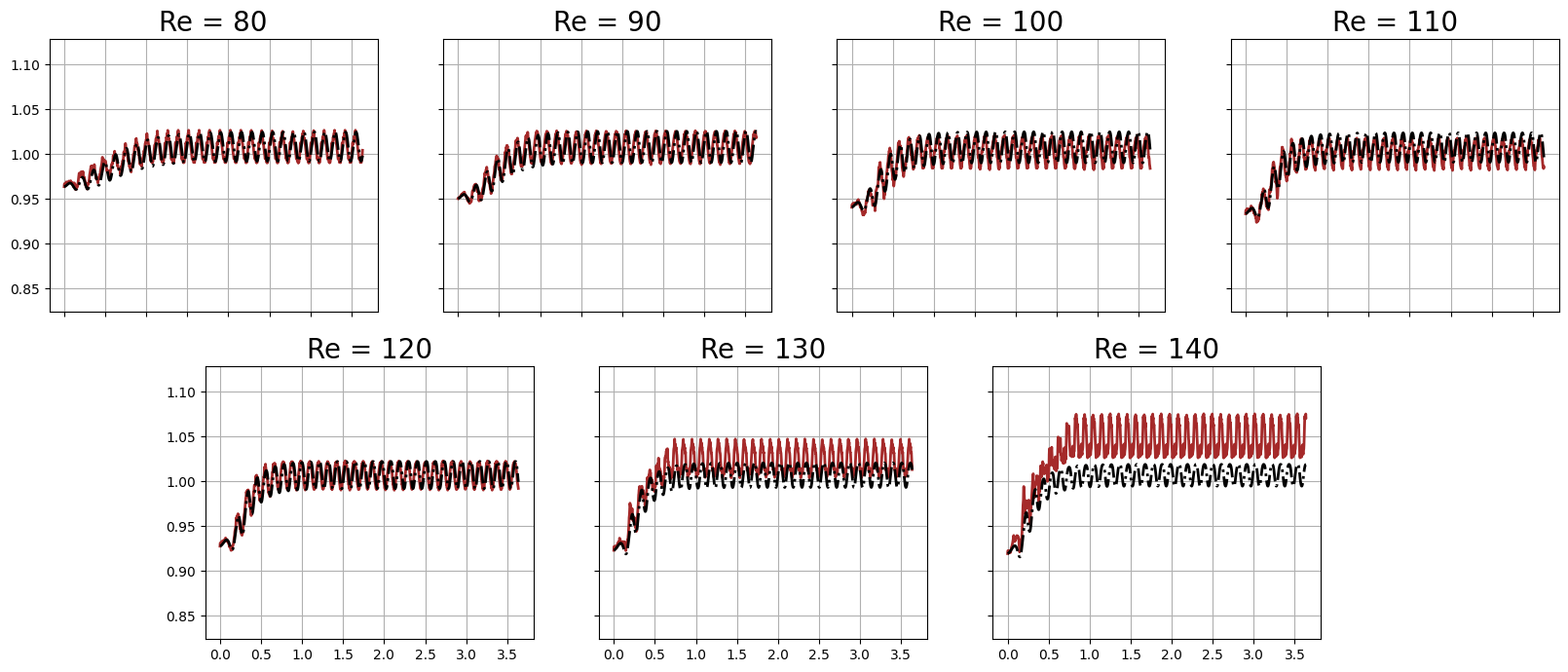}
    \caption{Kinetic energy prediction vs ground truth (validation set).}
    \includegraphics[width=0.3\linewidth]{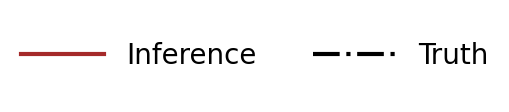}
    \label{fig:k_pred}
\end{figure}

\begin{figure}[!htb]
    \centering
    (a)\hfill (b)\hfill $\,$
    \includegraphics[width = \textwidth]{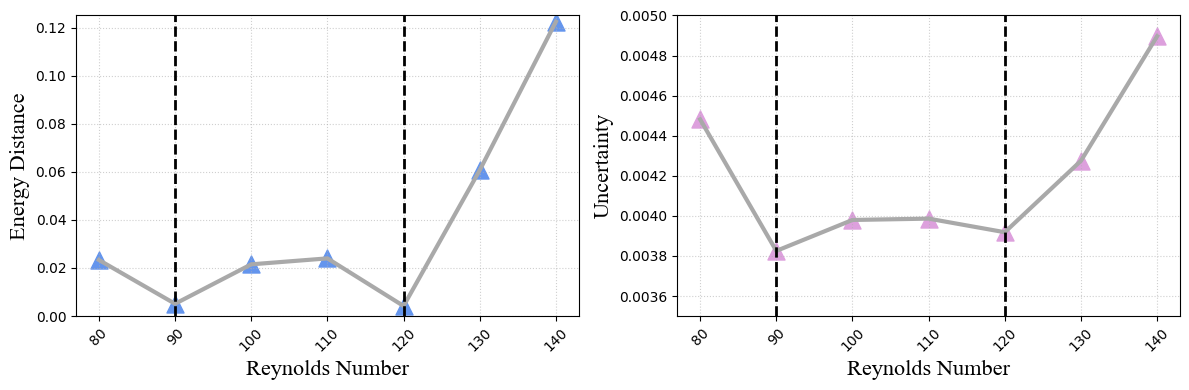}
    \caption{Comparison between (a) energy distance and (b) uncertainty quantification for different Reynolds numbers. The training samples ($Re = \{90, 120\}$) are indicated with the vertical dashed lines. }
    \label{fig:UQ_WD_init}
\end{figure}

Figure \ref{fig:k_pred} shows the ROM's predicted kinetic energy for all the validation cases.  
After training, the model is given only the initial condition and the associated Reynolds value, and uses auto-regression to forecast the entire dynamical window. %It is possible to qualitatively assess the model's performance by visualizing the kinetic energy signal spatially aggregated resulting in a $\mathbb{R}^{7, T/2}$ signals for the 7 Reynolds parameters of the validation set. 
Qualitatively, we observe that the model performs well within the training distribution ($Re = \{90, 120\}$). The interpolation regime yields overall satisfactory forecasting performance. However, extrapolation beyond $Re > 120$ deviates significantly from the reference data. 
This behaviour can be quantitively verified \textit{a posteriori} using the energy distance between the predicted and numerically generated data (Figure~\ref{fig:UQ_WD_init}a). 
Importantly, we show that the \textit{a priori} model’s UQ prediction (Figure~\ref{fig:UQ_WD_init}b) shows a good agreement with the energy distance, which means that we can estimate the performance of the model in real time, bypassing the need for \textit{a posteriori} analysis. The present analysis builds upon the UPdROM framework introduced in~\citep{UPdROM}, where the relationship between prediction error and the associated uncertainty is formally investigated. In particular, the study reports calibration metrics such as the Continuous Ranked Probability Score (CRPS), and evaluates the Pearson correlation coefficient between the scalar total prediction error and the corresponding scalar total uncertainty. The reported results demonstrate a strong linear relationship between these quantities, supporting the reliability of the proposed uncertainty estimates. The reader is referred to~\citep{UPdROM} for a more detailed quantitative validation of the uncertainty quantification framework.

The most uncertain and least accurate inference is observed at $Re = 140$, i.e., the extrapolation case. This behaviour is commonly seen in data-driven and machine learning models. The following study therefore focuses on improving the model’s performance at this Reynolds number while minimizing both the training and data costs.\\

\begin{figure}[!htb]
    \centering
    \includegraphics[width=\linewidth]{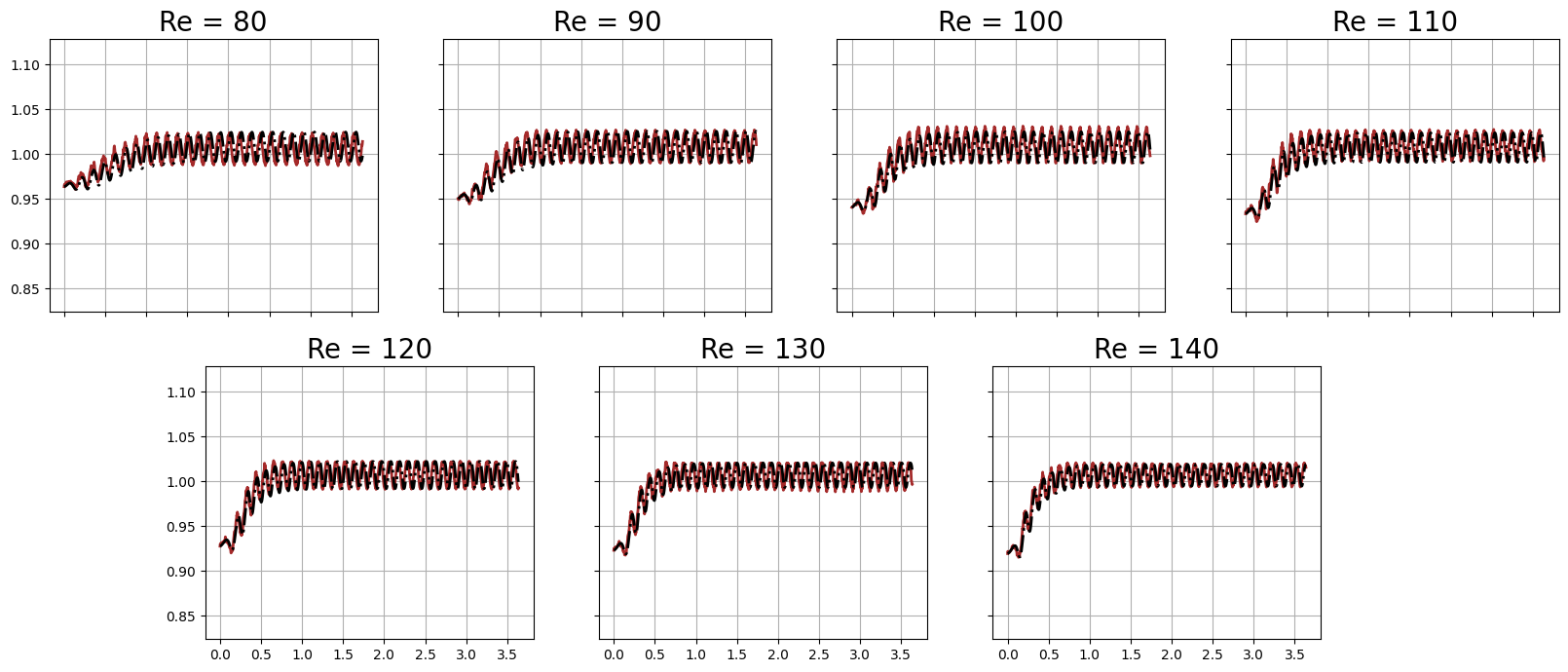}
    \caption{Kinetic energy prediction vs ground truth (validation set) after retraining of the full model}
    \includegraphics[width=0.3\linewidth]{Images/Part1/legend_K_prediction_init.png}
    \label{fig:k_pred_fs}
\end{figure}

\begin{figure}[!htb]
    \centering
    (a)\hfill (b)\hfill $\,$
    \includegraphics[width = \textwidth]{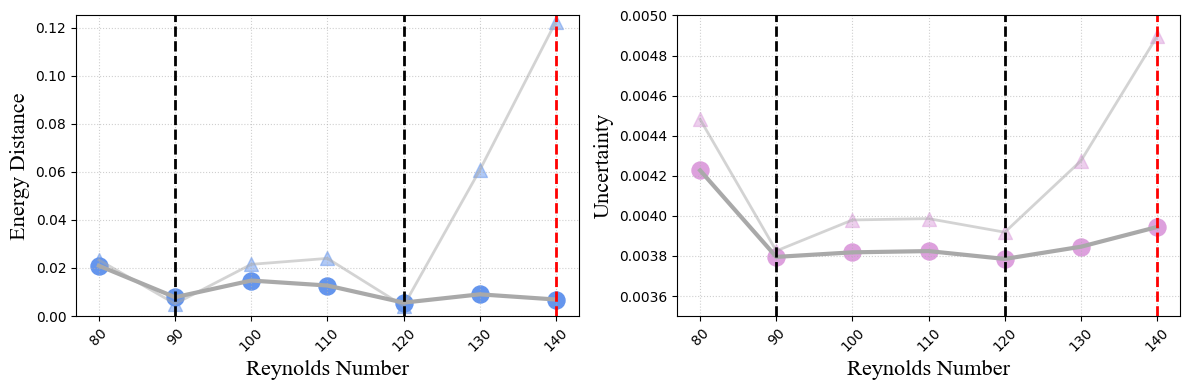}
    \includegraphics[width = 0.7\textwidth]{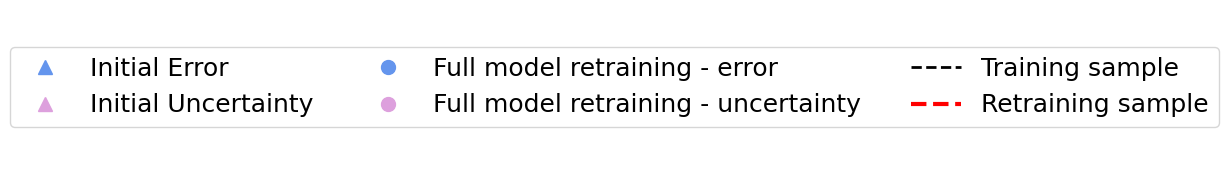}
    \caption{Comparison between energy distance and uncertainty quantification for different Reynolds numbers after retraining of the full model.}
    \label{fig:UQ_WD_fs}
\end{figure}

Classical methodology \citep{UPdROM} consists in finding this least performing point in the parametric space and retrain the model with true numerical data sampled in this learning bottleneck.  Essentially, a new simulation must be run at the selected parameter value, and the model must be fully retrained. This approach is standard in many surrogate model adaptive learning methods, which rely on high-quality data for adaptive training. Doing so provides a ``reference'' improvement, albeit at a relatively high retraining cost. 
This qualitative and quantitative improvement are illustrated via the the kinetic energy in Figure~\ref{fig:k_pred_fs}, and the \textit{a posteriori} energy distance and \textit{a priori} UQ in Figure~\ref{fig:UQ_WD_fs}.
Once again, a good agreement is observed between the UQ and the error. As expected, both the uncertainty and the error decrease substantially at the retraining sample, which is consistent with the kinetic energy signals at $Re = 140$ now closely matching those of the ground truth. Overall, the error and uncertainty are reduced over the entire parameter range, and the inference performance at $Re = 140$ approaches its upper performance bound. The objective is therefore to evaluate how closely this bound can be approached using more cost-effective methodologies.
\section{Manifold analysis}
\label{sec:manifold}
For the dynamical system under consideration, we hypothesise that the degradation of out-of-sample predictive accuracy is primarily caused by shifts in the learned low-dimensional manifold, rather than by inaccuracies in the inferred dynamics defined on that manifold. When the parameterised model is trained within the identified range of Reynolds numbers, the governing dynamics in latent space remain largely consistent for nearby Reynolds numbers. This implies that fine-tuning only needs to adjust the manifold itself, leaving the latent dynamics essentially intact. In practical terms, this suggests that fine-tuning the VAE while keeping the Transformer frozen, yield results as accurate, as a full-model retraining. By “fixing the manifold,” we refer specifically to adapting its geometry through targeted fine-tuning, since this geometric structure is the component most sensitive to variations in the Reynolds number. In practice, we freeze the Transformer weights and adapt only the VAE, while evaluating the model using the same, initial loss function that includes the penalty on the latent-space dynamical rollout error. This strategy allows the VAE to preserve favourable dynamical properties in the latent representation.  We now adopt a terminology to facilitate the distinction between the two adaptation strategies. We refer as 'retraining' the adaptation of the entire model, transformer included. We refer as 'fine-tuning', the VAE adaptation.

\begin{figure}[!htb]
\centering

% -------- Row 1 --------

\begin{subfigure}{0.80\linewidth}
    \centering
    \includegraphics[width=\linewidth]{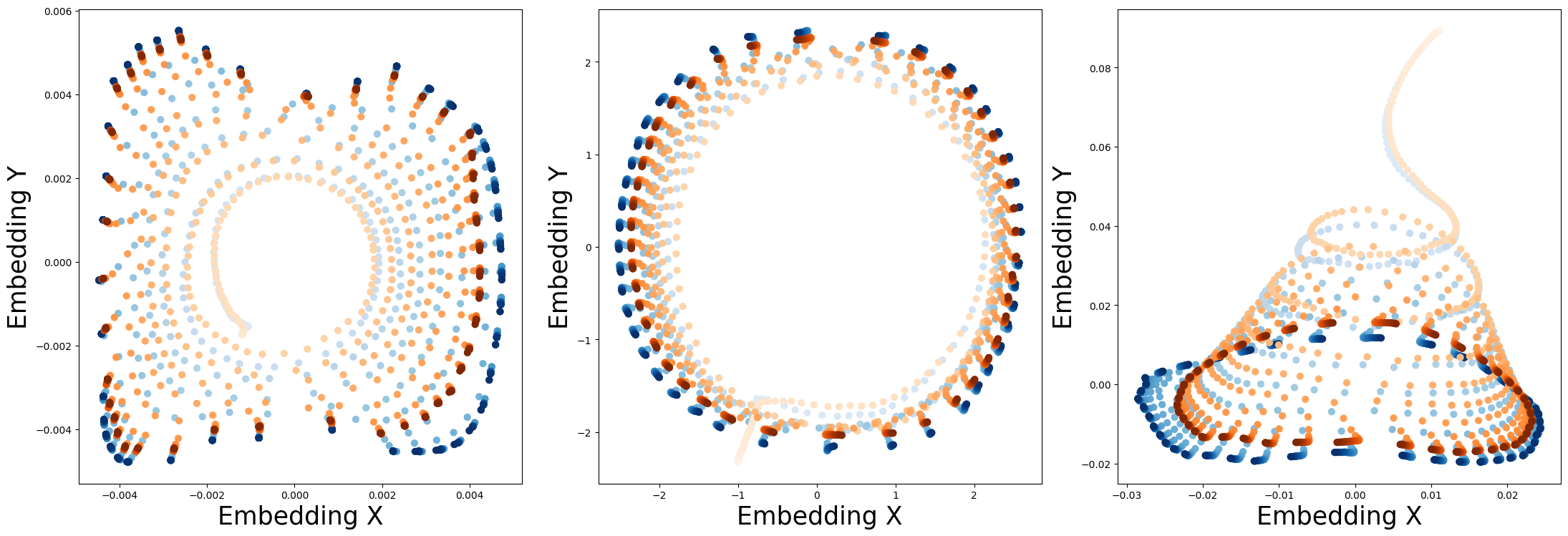}
    \caption{Spectral, Isomap and Hessian embeddings of the latent trajectories before \textit{(blue)} and after retraining \textit{(orange)}.}
    \label{fig:embeddings_fullretrain}
\end{subfigure}

\vspace{1em}

\begin{subfigure}{0.80\linewidth}
    \centering
    \includegraphics[width=\linewidth]{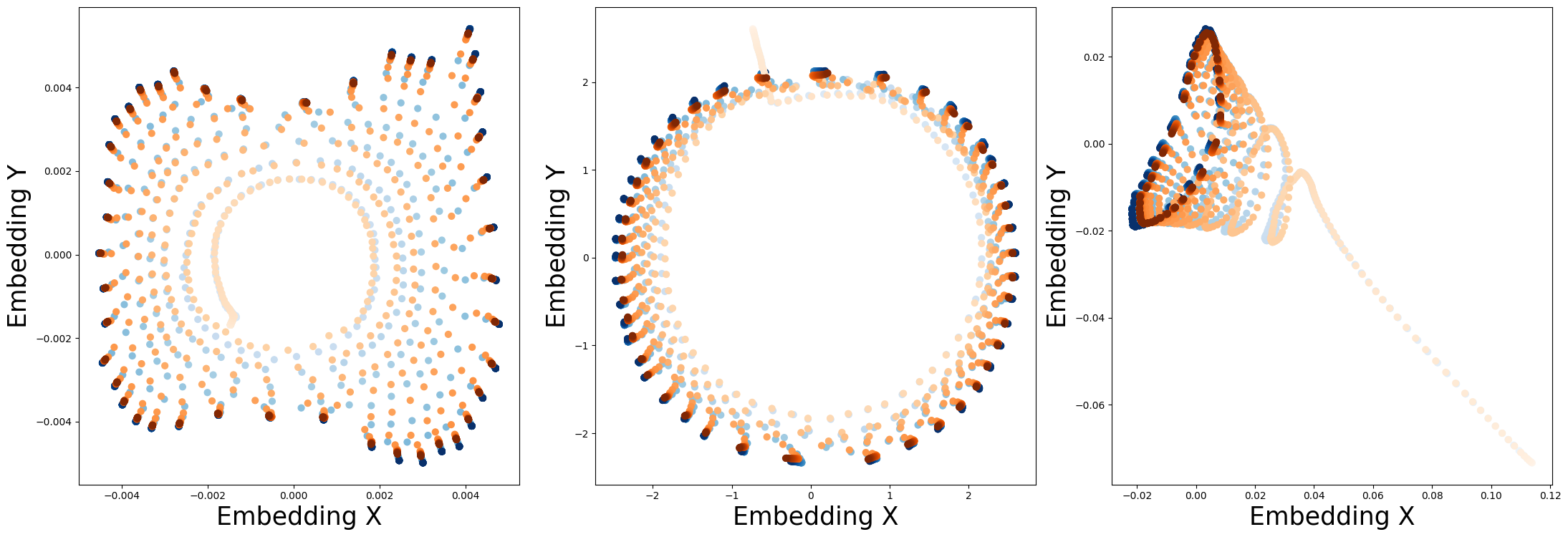}
    \caption{Spectral, Isomap and Hessian embeddings of the latent trajectories after VAE fine-tuning} \textit{(blue)} and after full model retraining \textit{(orange)}.
    \label{fig:embeddings_VAEretrain}
\end{subfigure}

\caption{Latent trajectories before retraining \textit{(blue, top)}, after VAE fine-tuning} \textit{(blue, bottom)}, and after complete retraining \textit{(orange)}. The time progression along each trajectory is indicated by an increasing colour intensity.

\end{figure}

We now demonstrate this hypothesis empirically. Among the evaluated cases, inference at \( Re = 140 \) was found to be the least confident and the least accurate. Full model retraining showed that the energy distance could be reduced by 93\%. We suggest that the projection onto the low-dimensional space is the most error-prone component of the model. This can be formally assessed by comparing the performance of the VAE before and after retraining, using validation data. In this analysis, we isolate the VAE from the full model by evaluating only its ability to reconstruct input data, without involving the temporal dynamics. Whenever dynamics is not involved, the distance metrics can be euclidean as no temporal shift may arise. The task at which the VAE is evaluated is pixel-to-pixel reconstruction. Table \ref{tab:metrics_MSE_VAE} compares this metric before and after retraining at $Re=140$.  
\begin{table}[!htb]
\centering
\caption{RMSE (in \%) before and after VAE retraining.}
\begin{tabular}{lcc}
\hline
\textbf{Metric} & \textbf{Pre-retraining} & \textbf{Post-retraining} \\
\hline
Relative \( L_1 \) error & 2.53\% & 0.19\% \\
Relative \( L_2 \) error & 3.11\% & 0.35\% \\
\hline
\end{tabular}
\label{tab:metrics_MSE_VAE}
\end{table}
This significant reduction in pixel-wise reconstruction error indicates that the VAE, prior to retraining, was indeed a limiting factor in the model’s overall performance. The improved accuracy following retraining suggests that many of the discrepancies observed in the full model's predictions may be attributed to deficiencies in the initial VAE’s ability to represent the flow fields accurately. This error can be attributed to the fact that the non-retrained VAE fails to infer the correct manifold, instead constructing one with inaccurate geometry. Even if the model learns the correct underlying dynamics, those dynamics will evolve incorrectly if they are rolled out on a flawed latent manifold. Figure~\ref{fig:embeddings_fullretrain} illustrates the latent trajectories generated by the model before and after full retraining. To provide a qualitative view of the underlying manifold topology, these trajectories are projected into two dimensions using nonlinear dimensionality reduction techniques, specifically, Spectral Embedding, Isomap, and Hessian Eigenmaps. The resulting visualizations reveal a geometric mismatch between the two latent trajectories.

% \begin{figure}[!htb]
% \centering
% \includegraphics[width=0.8\linewidth]{Images/Part2/latent_space_prepost.pdf}
% \vspace{1em}
% \includegraphics[width = 0.3\linewidth]{Images/Part2/legend_manifold_fs.png}
% \caption{Latent trajectories before \textit{(blue)} and after retraining \textit{(orange)}.}
% \label{fig:latent_manifold_prepost}
% \end{figure}

% \begin{figure}[!htb]
% \centering
% \includegraphics[width=0.8\linewidth]{Images/Part2/Embeddings_full_retrain.png}
% \vspace{1em}
% \includegraphics[width = 0.3\linewidth]{Images/Part2/legend_manifold_fs.png}
% \caption{Spectral, Isomap and Hessian embeddings of the latent trajectories}
% \label{fig:embeddings_fullretrain}
% \end{figure}
The embeddings suggest that the most significant mismatches occur along the periodic orbits. The radius of the predicted periodic orbit in the latent space is larger in the non-retrained model, indicating a flawed reconstruction of the system’s limit cycle. This discrepancy is consistent with the kinetic energy signal predicted by the non-retrained model at $Re=140$, where the amplitude of oscillations deviates significantly from the ground truth.
We now show that fine-tuning the VAE fixes the topology of the latent manifold and consequently improves the error just as well as a complete retraining. 

Completing the fine-tuning solely onto the VAE, Figure \ref{fig:embeddings_VAEretrain} shows the topology of the manifold in 2D embedding spaces. 

% \begin{figure}[!htb]
% \centering
% \includegraphics[width=0.8\linewidth]{Images/Part2/latent_space_postVAEpost.pdf}
% \caption{Latent trajectories after VAE retraining \textit{(blue)} and after full model retraining \textit{(orange)}.}
% \label{fig:latent_manifold_postVAEpost}
% \end{figure}

% \begin{figure}[!htb]
% \centering
% \includegraphics[width=0.8\linewidth]{Images/Part2/Embeddings_VAEretrain.png}
% \caption{Spectral, Isomap and Hessian embeddings of the latent trajectories}
% \vspace{1em}
% \includegraphics[width = 0.5\linewidth]{Images/Part2/legend_latent_traj_VAE.png}
% \label{fig:embeddings_VAEretrain}
% \end{figure}

\begin{figure}[!htb]
    \centering
    \includegraphics[width = \textwidth]{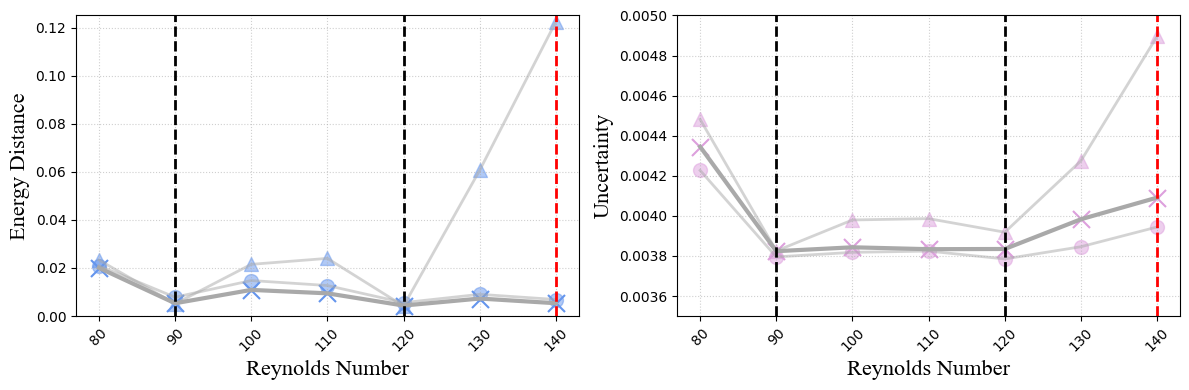}
    \includegraphics[width = \textwidth]{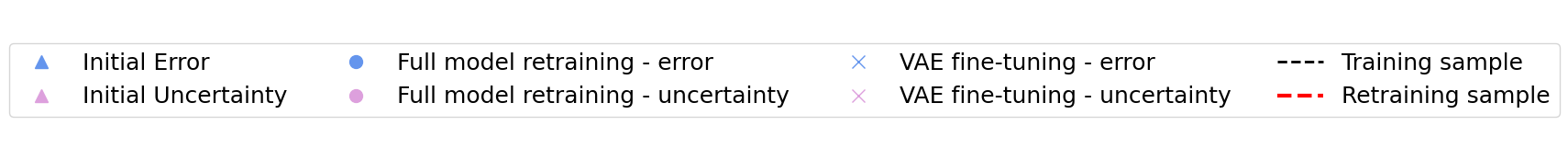}
    \caption{Comparison between energy distance and uncertainty quantification for different Reynolds numbers after fine-tuning} of the VAE.
    \label{fig:UQ_WD_VAE}
\end{figure}

Fine-tuning the VAE fixes the projection operator just as well as a complete retraining as exhibited by table \ref{tab: L1-L2 reconstruction error of VAE}
\begin{table}[!htb]
    \centering
\begin{tabular}{lccc}

    \textbf{Metric} & \textbf{Pre-retraining} & \textbf{Post complete retraining} & \textbf{Post-VAE fine-tuning} \\
    \hline
    Relative \( L_1 \) error & 2.53\% & 0.19\% & 0.19\% \\
    Relative \( L_2 \) error & 3.11\% & 0.35\% & 0.37\% \\
    \hline

\end{tabular}    
\caption{ L1-L2 reconstruction error of VAE}
\label{tab: L1-L2 reconstruction error of VAE}
\end{table}
Reconsidering now the complete dynamical model, with the Transformer operating in the latent space, we evaluate the model uncertainty and the energy distance across the range of Reynolds numbers. The results of this analysis are presented in Figure~\ref{fig:UQ_WD_VAE}.

The energy distance remains as low as in the case of complete retraining, demonstrating that the Transformer does not need to relearn the underlying dynamics as long as the latent manifold accurately captures the essential modal structure of the flow. This result is significant: the time-dependent behaviour of the system remains consistent, while the structure of the observable space is effectively reshaped through fine-tuning. The first key finding of this study is that the cost required for fine-tuning the model is considerably less extensive than initially anticipated, thereby necessitating reduced computational time and a smaller amount of data. Figure \ref{fig:bar_plot_error} provides a quantitative comparison, supporting the advantage of solely adapting the VAE. In our case, fine-tuning on an NVIDIA RTX 4000 Ada Generation GPU required less than a minute to match the first-moment statistics, enabling near-real-time adaptation. Obtaining meaningful second-moment statistics, however, typically takes longer, around 15 minutes, compared to the 2 hours required to retrain the entire model. This efficiency holds as long as the system does not encounter a bifurcation, beyond which the flow would exhibit qualitatively distinct behaviour. Building on this cost-effective approach, the next section demonstrates that fine-tuning can also rely on highly sparse observations, further reducing both data and computational demands.
% The \ref{subsec:transfoRetrain} shows a similar methodoly applied to the Transformer, where the VAE is kept frozen at the retraining stage. 

\section{Data Assimilation with sparse observations}
\label{sec:DA}

In this section, we can assess whether we can improve the ROM's predictions from sparse observations using sequential data assimilation.  
With this, we aim to reduce further the overall data requirements and providing a pathway toward  near-real-time model correction using available numerical/experimental observations.
Specifically, we take an ensemble Kalman filter (EnKF) approach, which  enables real-time correction of the model using limited measurements~\citep{Evensen2003, Bocquet, Real_time_calibration_chaotic}. 
%
% The motivation to use the EnKF is three-fold: 
% (i) the ROM predictions exhibit approximately Gaussian statistics, (ii) the system dynamics are nonlinear and high-dimensional, and 
% (iii) the EnKF is a non-intrusive computationally efficient assimilation.

\subsection{Iterative Model Refinement via Data Assimilation}
\label{subsec:DA_framework}

The proposed ROM, here denoted $\tilde{g}$, produces a probabilistic output by design. 
As discussed in Sec.~\ref{subsec:ROM}, the ROM error can be represented by an uncertainty estimate derived from a latent Gaussian distribution. 
The nonlinear nature of the decoder does not theoretically guarantee that this Gaussianity is preserved in the physical output space. However, we perform a two-sided Kolmogorov–Smirnov test, which  indicates that the null hypothesis (i.e., that the decoded ensemble follows a Gaussian distribution) cannot be rejected.  The test was conducted independently for each snapshot-location pair, and showed that $99.04\%$ of the tested pairs failed to reject the Gaussianity null hypothesis. The test was performed on the generated forecast ensemble $\psi^f$, which has a limited cardinality ($N=20$). Consequently, the quantitative results must be interpreted cautiously due to the limited statistical power associated with small sample sizes. To complement this analysis, we provide Q-Q plots in \ref{subsec:gaussianity} with figure \ref{fig:qqplot} in order to qualitatively assess the Gaussianity assumption.
By confirming that the ROM generally satisfies the assumption of Gaussian model error, we can reliably employ the EnKF to combine the ROM output state with sparse observations in a mathematically consistent manner. This is, we can employ an ensemble to accurately represent the statistics of the prediction via an ensemble approximation.

We sample an ensemble of states from the latent space as shown in Figure~\ref{fig:UQ}, which is forecast in time in the latent space. 
This ensemble is denoted by $\psi^f$, where the superscript $f$ stands for \textit{forecast}, and the ensemble cardinality is $|\psi^f| = N$,  $\psi^f_{i\in N} \in \mathbb{R}^{\text{T} \times \text{m}}$ where \textit{T} is the time-wise dimension, and  the spatial dimension is large  $m\sim O(10^4)$. 
The best estimate of the ROM output and the uncertainty are approximated by the ensemble mean  $X^f_t$ and covariance  $P^f_t$, i.e., 
\begin{equation}\label{eq:ensemble}
X^f = \frac{1}{N} \sum_{i=1}^N \psi^f_i
\quad \text{and} \quad
P^f_t = \frac{1}{N - 1} \sum_{i=1}^N (\psi^f_{i,t} - X^f_t)^T(\psi^f_{i,t} - X^f_t). 
\end{equation}

To improve the forecast ensemble, we assimilate sparse observation data, $Y_{\text{obs}}$, which represent partial measurements from a limited number of sensors positioned within the flow domain such that
\[
Y_{\text{obs}} = X_{\text{true}} H^T. 
\]
where $X_{\text{true}}$ is the true state data of dimension $(T, m)$, and $H$ is a downsampling operator: $ H \in \mathbb{R}^{\text{nobs} \times \text{m}}$. The matrix $H$ is sparse, containing $n_{\text{obs}} \ll m$ ones and zeros elsewhere, effectively selecting a subset of the state variables. Consequently, $Y_{\text{obs}} \in \mathbb{R}^{\text{T}\times \text{nobs}}$, does not represent a full-state observation but rather partial measurements, such as those obtained from a limited number of sensors positioned within the flow domain. 
We use the observation operator to obtain the ROM's estimate on the observation points is  $Y^f_{j} =   \psi^f H^T$ for $i \in N$. 
In addition, we define the sparse forecast cross-covariance $P^fH^T$, also referred to as the \textit{influence function} \cite{Evensen2003} 
\begin{equation}
\label{eq:Influence_function}
    P^f H^\top = \frac{1}{N - 1} \sum_{i=1}^N (\psi^f_{t,i} - X^f_t)^\top (\psi^f_{t,i} H^\top - X^f_t H^\top). 
\end{equation} 
With this, we bypass the need to compute and store the full covariance matrix  $P^f_t \in \mathbb{R}^{\text{m} \times \text{m}}$ \eqref{eq:ensemble} at each time step, which poses a significant computational challenges due to its high dimensionality.  \\

The EnKF update (or analysis step) at each time step $t$ reads 
\begin{equation}
    \psi^a_t = \psi^f_t + K (Y_{\text{obs},t} - H \psi^f_t)
\end{equation} 
where $a$ stands for \textit{analysis}, 
the Kalman gain $K = P^f H^\top (HP^fH^\top + R)^{-1}$, 
and $R$ is the observation error covariance matrix. 
 It is defined with an \textit{artificial} noise level \(\epsilon\), since the observations considered in this study are synthetic and noise-free, being directly sampled from the ground-truth solution through the subsampling operator \(H\). Consequently, \(\epsilon\) is introduced purely for numerical stability purposes and is set to \(10^{-7}\). This value remains negligible compared to the typical magnitude of the velocity variables, thereby preserving near-perfect confidence in the observations while ensuring a stable assimilation procedure. The results were found to be insensitive to small variations of \(\epsilon\) within the same order of magnitude.
The resulting analysis ensemble, $\psi^a_t$, provides both an optimal state estimate via its mean, $X^a_t$, and a measure of uncertainty via its covariance  $P^a_t$.  
This newly available full-state data can be used to retrain the model, using the following loss function to improve the model forecast 
\begin{equation}\label{eq:loss}
    \left\lVert \tilde{g}(X_{t-1}) - X^a_t \right\rVert_2. 
\end{equation}
% Here, \( X^a \) denotes the mean of the analysis ensemble, defined as
% \[
% X^a = \frac{1}{N} \sum_{i=1}^N \psi^a_i \in \mathbb{R}^{m}.
% \]
Previous studies that combine assimilated data with machine learning have proposed alternative loss functions that explicitly account for the inherent uncertainty of the Kalman filter. For instance, \citep{bocquet_2} introduced a weighted loss that penalizes confident errors more heavily than uncertain ones with the cost function
\begin{equation}
\label{eq:alternative_loss}
    \left\lVert \tilde{g}(X_{t-1}) - X^a_t \right\rVert_{2, P_k^{-1}},
    \quad \text{where} \quad 
    \left\lVert x \right\rVert_{2, P_k^{-1}} = x^\top P_k^{-1} x.
\end{equation}
In this formulation, \( P_k \) acts as the surrogate model error covariance matrix, so that more uncertain states (with larger variance) receive smaller weights in the loss. However, this approach requires computing the inverse of a high-dimensional covariance matrix at each training step, which incurs a prohibitive computational cost. Furthermore, Section~\ref{subsec:second_moment} illustrates that, due to the architecture of our model, there is no need to account for aleatoric uncertainties arising from the Kalman filter, meaning that it does not need to be embedded in the loss through the computation of the error covariance. 
Therefore, we adopt the simpler quadratic distance given in  Equation~\ref{eq:loss}.  No covariance inflation, localization, or additional regularization strategies were employed in the present study.
 
Algorithm \ref{alg:hybrid_training} summarizes the procedure followed for iterative ROM refinement from sparse data.   
\begin{algorithm}[!ht]
\caption{Iterative Model Refinement via EnKF Assimilation}
\label{alg:hybrid_training}
\begin{algorithmic}[1]

\Require Training distribution $\Omega_{train} \in \mathbb{R}^{T \times m}$, Initial model $\tilde{g}$, Inference parameter $\xi \notin \Omega_{train}$, Initial condition $X_0 \in \mathbb{R}^{T \times m} $
\Ensure Updated model $\tilde{g}$

\State \textbf{Step 1: Ensemble Forecasting}
\State Generate forecast ensemble $\tilde{g} : X_0, \xi \rightarrow \Psi^f_{\xi} \in \mathbb{R}^{N \times T \times m} $

\State \textbf{Step 2: Ensemble Kalman Filter (EnKF)}
\For{$t = 1$ \textbf{to} $T$}
    \State Extract ensemble $\psi^f_t \in \mathbb{R}^{N \times m}$
    \State Compute the ensemble mean $X^f_t = \frac{1}{N} \sum_{i=1}^N \psi^f_{t,i} \in \mathbb{R}^{m}$
    \State Compute sparse covariance matrix $P^f H^\top \in \mathbb{R}^{m \times n_{\text{obs}}}$ via Eq. \ref{eq:Influence_function}
    \State Compute projected covariance $HP^fH^\top \in \mathbb{R}^{n_{\text{obs}} \times n_{\text{obs}}}$
    \State Define observation noise covariance $R = \epsilon \mathbb{I}_{n_{\text{obs}}}, \quad \epsilon \ll 1$
    \State Compute Kalman gain $K \in \mathbb{R}^{m \times n_{\text{obs}}}$:
    $$K = P^f H^\top (HP^fH^\top + R)^{-1}$$
    \State Obtain the analysis ensemble $\psi^a_t \in \mathbb{R}^{N \times m}$:
    $$\psi^a_t = \psi^f_t + K (Y_{\text{obs},t} - H \psi^f_t)$$
\EndFor

\State \textbf{Step 3: Model Retraining}
\State $\Omega_{train} \leftarrow \Omega_{train}\cup X^a $  with $X^a = \frac{1}{N} \sum_{i=1}^N \psi^a_i \in \mathbb{R}^{T \times m}$.
\State Retrain model $\tilde{g}$ by minimizing the loss:
$$\mathcal{L} = \left\lVert \tilde{g}(X_{t-1}) - X^a_t \right\rVert_2$$

\State \Return $\tilde{g}$

\end{algorithmic}
\end{algorithm}

\subsection{State estimation results}

\begin{figure}[H]
    \centering
    \begin{subfigure}[t]{0.35\textwidth}
        \includegraphics[width=1\linewidth]{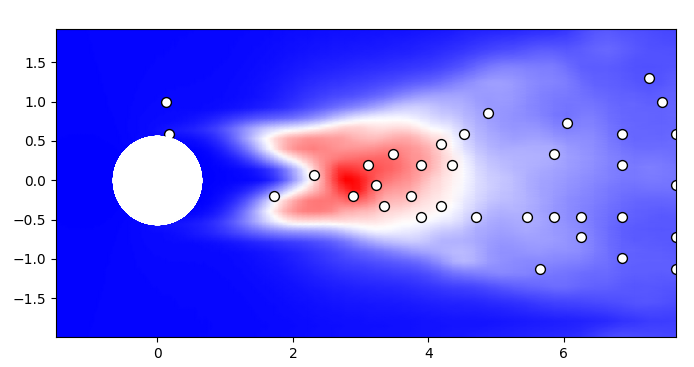}
        \caption{Sensor locations for $U$ velocity field, represented on the average U uncertainty}
        \label{fig:sensors_U}
    \end{subfigure}
    \hspace{2em}
    \begin{subfigure}[t]{0.35\textwidth}
        \includegraphics[width=1\linewidth]{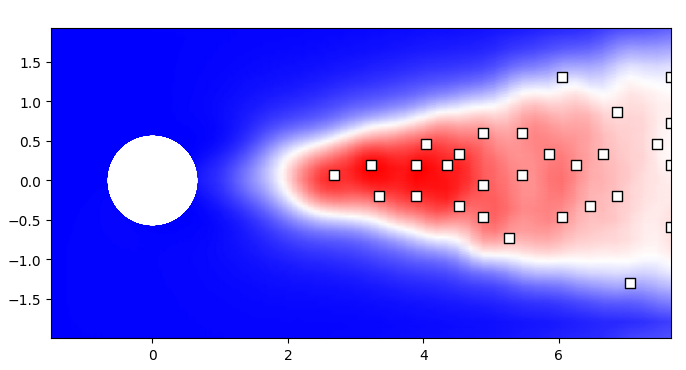}
        \caption{Sensor locations for $V$ velocity field, represented on the average V uncertainty}
        \label{fig:sensors_V}
    \end{subfigure}
    \begin{subfigure}[t]{0.12\textwidth}
        \includegraphics[width=1\linewidth]{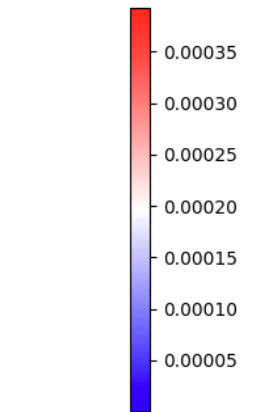}
    \end{subfigure}
    \caption{Sensor placement computed for both velocity fields represented upon their respective mean field.}
    \label{fig:sensors}
\end{figure}

We evaluate the state estimation performance for the flow at Reynolds number 140. 
To maximize the captured modal information, we position 64 sensors, distributed across the $U$ and $V$ velocity fields according to the criteria detailed in \ref{subsec:sensors}. These 64 high-fidelity observations represent only 1\% of the total spatial degrees of freedom ($m$).
Figure~\ref{fig:sensors} shows the location of the sensors overlaid on the mean uncertainty of both velocity fields. Logically, they are positioned downstream of the obstacle and in the wake of the uncertainty. By concentrating measurements in these error-prone regions, we maximize the information gain provided to the EnKF.

 The forecast ensemble \(\psi^f\) is generated with a cardinality of \(N = 20\). This choice was primarily motivated by the goal of maintaining a lightweight and computationally inexpensive correction strategy. Furthermore, a short comparative study was conducted to evaluate the impact of both the number of sensors and the ensemble cardinality. The corresponding results, reported in \ref{subsec:sensors} with the table ~\ref{tab:cardinality_vs_RNMSE}, indicate that increasing the ensemble size to \(N = 100\) does not yield substantially improved assimilation performance.

\begin{figure}[H]
    \centering
    \includegraphics[width=1\linewidth]{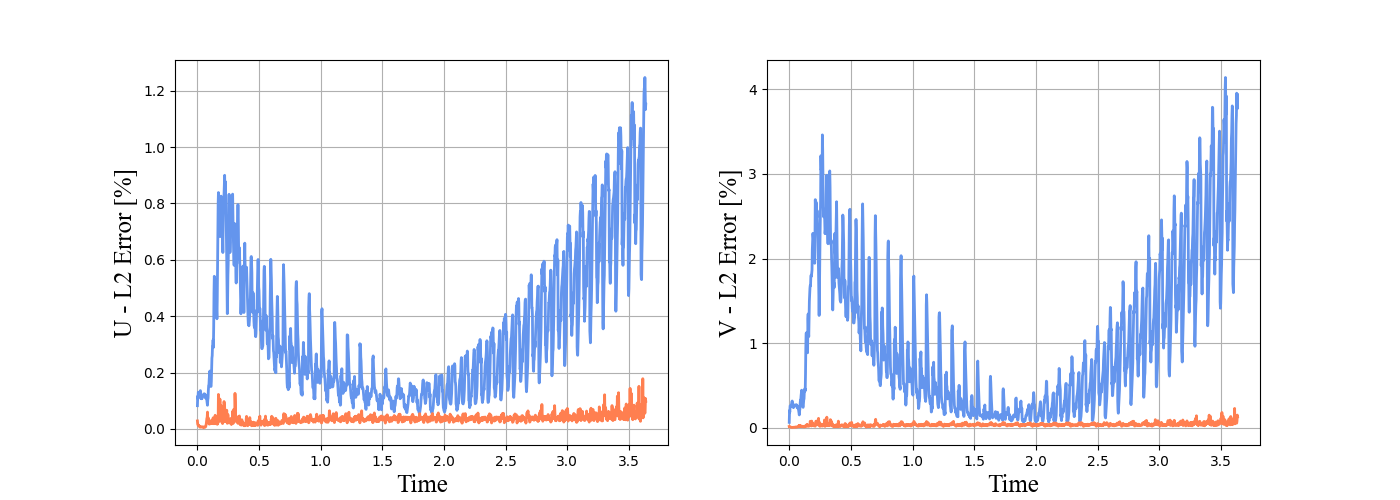}
    \caption{Forecast and filter error throughout time. }
    \includegraphics[width=0.6\linewidth]{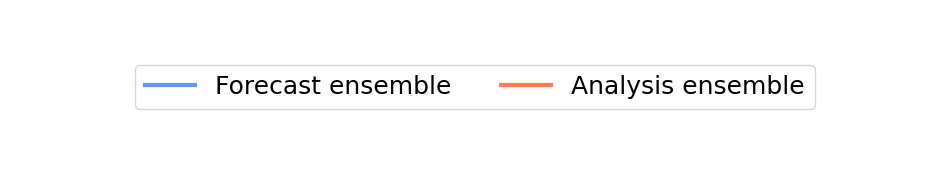}
    \label{fig:DA-time}
\end{figure}

\begin{figure}[H]
    \centering
    \includegraphics[width = 0.8\textwidth]{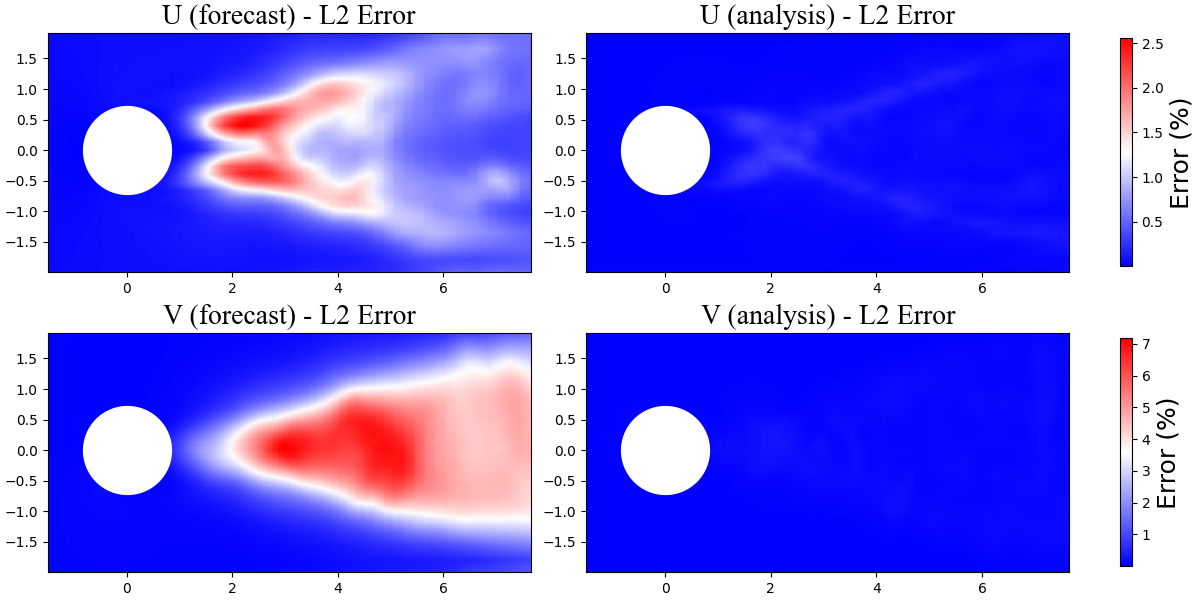}

    \caption{Error spatial distribution}
    \label{fig:Space_err}
\end{figure}

We apply the EnKF to obtain an analysis mean which is more accurate than the initial ROM estimate following the procedure outlined in Algorithm~\ref{alg:hybrid_training} (Steps 1–2). The prediction error is significantly reduced across both temporal and spatial domains, as illustrated in Figures~\ref{fig:DA-time} and \ref{fig:Space_err}, respectively. The error that is considered here is the quadratic distance since Kalman filter analysis minimizes a quadratic distance-based cost function. We quantify the performance of the Ensemble Kalman Filter through the relative reduction of the $\ell_2$ error. More specifically, this metric measures the improvement in the quadratic distance between the ground truth and the mean forecast ensemble $X^{f}$, compared to the quadratic distance between the ground truth and the mean analysis ensemble $X^{a}$.
Overall, the error in $U$ is reduced by $85.65\% \pm 0.59$, and the error in $94.29\% \pm 0.19$. Since the initial error in $V$ was larger than in $U$, the combined improvement across the two fields corresponds to an overall reduction of $91.34\% \pm 0.31$.
The phase and frequency errors introduced by the ROM, which produce the time-wise oscillatory behaviour in the error signal, are essentially eliminated after filtering. These oscillations occur at multiple scales because the ROM slightly mispredicts the frequencies associated with different temporal modes. Although these discrepancies are small, they accumulate in time until the mismatch catches up with the next period, at which point the error decreases. After filtering on the entire time horizon, these effects are removed. However, the analysis ensemble becomes noticeably noisier, which leads to the fuzzier error signals observed in the reconstructed fields, illustrated in figure \ref{fig:DA-time}. The assimilated data exhibit sufficient qualitative and statistical accuracy to be used for fine-tuning.

Specifically, we provide the model with data from a 64-dimensional observation space rather than a 6550-dimensional input space. The model leverages the estimates it is confident about and relies on the correlations between the observations and the less confident estimates. 

\subsection{Fine-tuning of the VAE with the analysis data}

Here, we showcase the Step 3 in Algorithm~\ref{alg:hybrid_training}, i.e., the fine-tuning of the VAE using analysis data.  
We observe a substantial improvement in both the \( L_1 \) and \( L_2 \) reconstruction errors in the projection to the latent space. Fine-tuning with assimilated data does not perform as well as fine-tuning with full-state numerical data, yet it offers a reasonable trade-off between performance and data requirements. Table~\ref{tab:L1-L2_error_DA} presents the comparative results.
\begin{table}[H]
    \centering
    \caption{Comparison of \( L_1 \)- and \( L_2 \)-norm reconstruction errors of the VAE across adaptation strategies. The acronym \textit{DA} stands for \textit{Data Assimilation}}
    \begin{tabular}{lccc}
        \textbf{Metric} & \textbf{Pre-retraining} & \textbf{Post full retraining} & \textbf{Post VAE fine-tuning} with DA \\
        \hline
        Relative \( L_1 \) error & 2.53\% & 0.19\% & 1.37\% \\
        Relative \( L_2 \) error & 3.11\% & 0.37\% & 1.98\% \\
        \hline
    \end{tabular}
    \label{tab:L1-L2_error_DA}
\end{table}

\begin{figure}[H]
\centering
\includegraphics[width=0.8\linewidth]{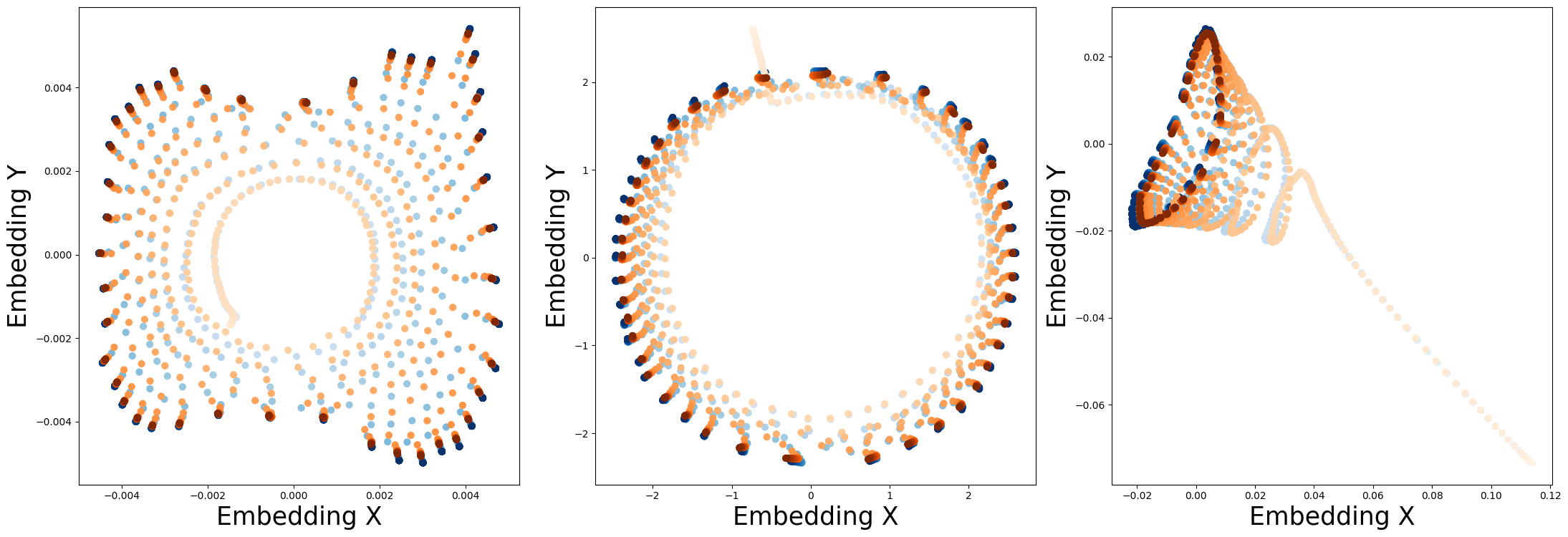}
\caption{Spectral, Isomap, and Hessian embeddings of the latent trajectories after fine-tuning} with assimilated data. The reference trajectory obtained after complete retraining is shown in \textit{orange}, while the trajectory obtained after fine-tuning the VAE with assimilated data is shown in \textit{blue}. The time progression along each trajectory is indicated by an increasing color intensity. 
\label{fig:embeddings_DAVAEretrain}
\end{figure}

Figure \ref{fig:embeddings_DAVAEretrain} shows that the latent manifold remains well structured when trained solely on assimilated data. Additionally, when compared with the best possible latent trajectory fit, the post-fine-tuning inference on assimilated data overlaps well, indicating that, geometrically and topologically, fine-tuning with DA performs comparably to retraining with full-state data. 

\begin{figure}[H]
    \centering
    (a)\hfill (b)\hfill $\,$
    \includegraphics[width = \textwidth]{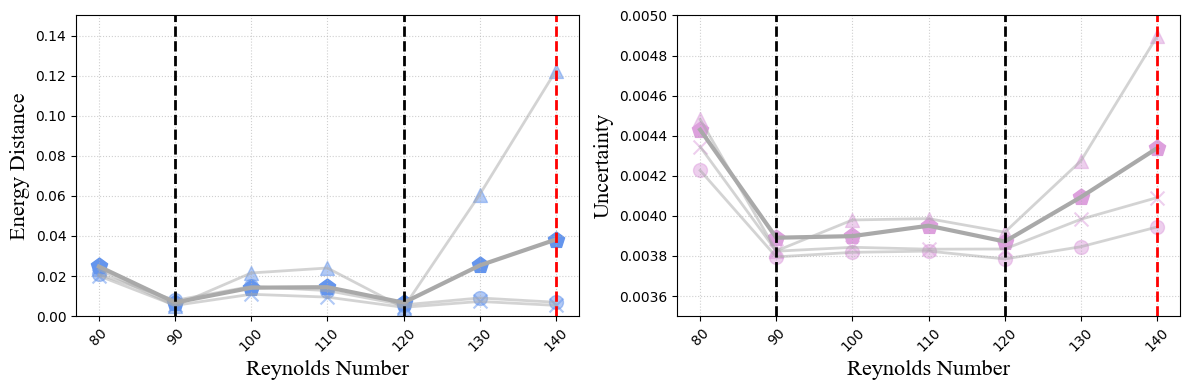}
    \includegraphics[width = 1\textwidth]{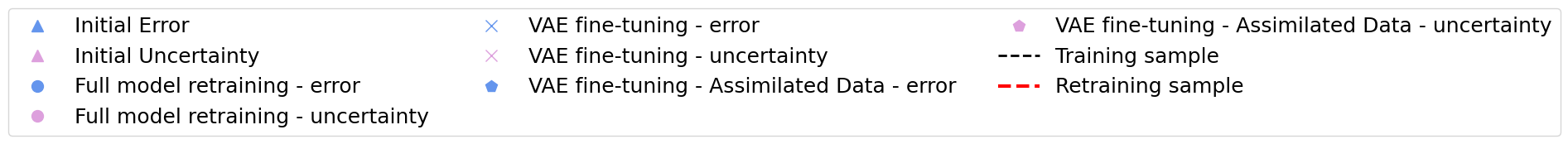}
    \caption{Comparison between (a) energy distance and (b) uncertainty quantification for different Reynolds numbers after VAE fine-tuning} with assimilated data.
    \label{fig:UQ_WD_DA}
\end{figure}

\begin{figure}[H]
    \centering
    \includegraphics[width=0.7\linewidth]{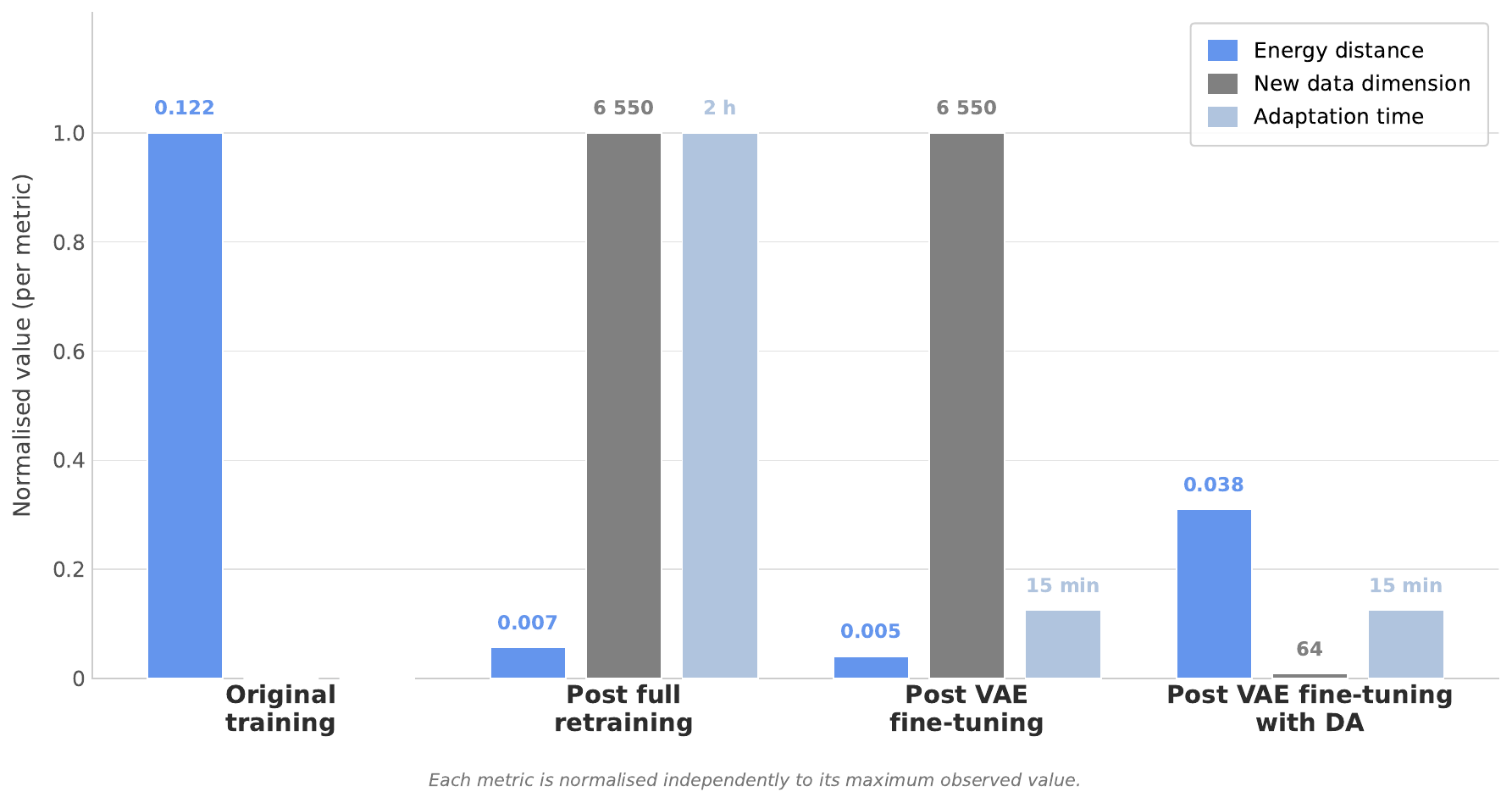}
    \caption{Comparison of energy distance at $Re=140$, the amount of new data required and time for each adaptation modality.}
    \label{fig:bar_plot_error}
\end{figure}

Considering the dynamical behaviour and model forecast quality, Figure~\ref{fig:UQ_WD_DA} shows the distribution of prediction error and uncertainty across the range of Reynolds numbers. The energy distance is significantly reduced compared to the pre-retraining stage, while using sparse observational data. The noisy signals produced by the EnKF data-assimilation step slightly increase the model’s overall predictive uncertainty. Although combining the assimilated data with the original training set reduces the risk of forgetting during fine-tuning, mixing data of different quality introduces a slight disturbance to how the model incorporates new information relative to what it has previously learned. Nevertheless, the impact remains minor and the model’s predictions remain largely robust.
Figure~\ref{fig:bar_plot_error} compares these results with other adaptation strategies discussed in Section~\ref{sec:manifold}. Fine-tuning the VAE using data assimilation offers a robust cost-performance balance and is particularly efficient, achieving comparable performance to full retraining with significantly reduced data and time requirements. 

\begin{figure}[H]
    \centering
    
    \begin{subfigure}[b]{0.48\textwidth}
        \centering
        \includegraphics[width=0.8\linewidth]{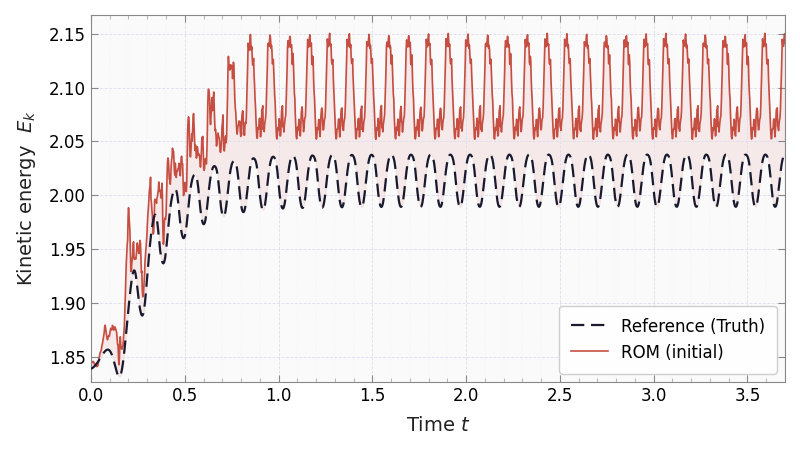}
        \caption{Kinetic energy forecast at $Re=140$ prior to fine-tuning.}
        \label{fig:k_before_ft}
    \end{subfigure}
    \hfill
    \begin{subfigure}[b]{0.48\textwidth}
        \centering
        \includegraphics[width=0.8\linewidth]{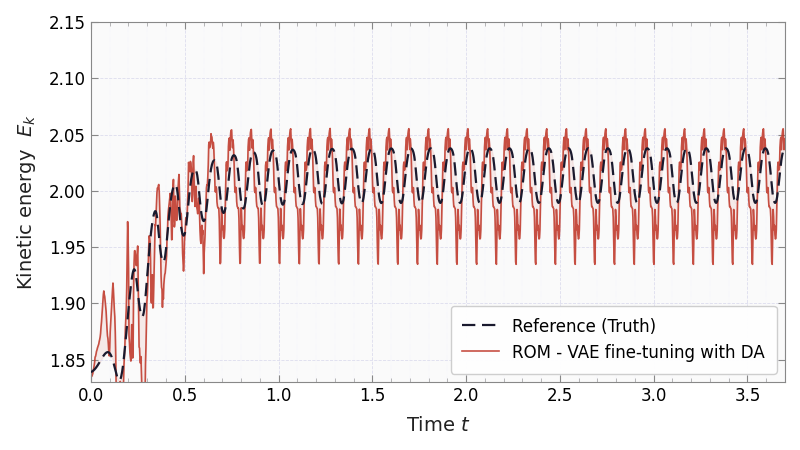}
        \caption{Kinetic energy forecast at $Re=140$ after 30 seconds of VAE fine-tuning using assimilated data.}
        \label{fig:k_after_ft}
    \end{subfigure}

    \caption{Comparison of the predicted kinetic energy evolution at $Re=140$ before and after rapid VAE fine-tuning with assimilated sparse-observation data.}
    \label{fig:k_DA_fast}
\end{figure}

We achieved nearly a 70\% reduction in energy distance while using only 1\% of actual new data at the end of the DA-fine-tuning pipeline. Model convergence is usually achieved after just a few training epochs, allowing for rapid, near-real-time adaptation. In practice, however, prediction of the first moment converges much earlier, within seconds, providing rapid fine-tuning capability when estimates of variance and uncertainty are not required. After just 30 seconds of fine-tuning using data assimilation, the model's error reaches its minimum. Figure \ref{fig:k_DA_fast} demonstrates this rapid correction, with the forecasted energy signal following the fine-tuning process. This result highlights the effectiveness and responsiveness of the approach, especially when only the model prediction needs updating. These results were obtained using a single GPU 'NVIDIA RTX 4000 Ada Generation'.

% \begin{figure} \centering \includegraphics[width=0.7\linewidth]{Images/Part3/kinetic_short_retraining.png} \includegraphics[width = 0.3\linewidth]{Images/Part1/legend_K_prediction_init.png} 
% \caption{Kinetic energy forecast after 30s of fine-tuning with DA} 
% \label{fig:k_DA_fast} 
% \end{figure} 

Regarding uncertainty quantification, the use of assimilated rather than full-state data has negligible impact on the uncertainty distribution across the flow field. Section~\ref{subsec:second_moment} provides a deeper analysis of the impact of DA on the uncertainty estimate and offers a justification for neglecting the additional uncertainty induced by the Kalman filter, measurable with the second moment of the analysis ensemble. Note that several alternative fine-tuning strategies could also have been considered in the present study for completeness. These include \textit{encoder-only} fine-tuning, \textit{decoder-only} fine-tuning, or partial layer freezing strategies in which only a subset of the autoencoder parameters is adapted while the remaining layers are kept frozen. Such approaches are commonly used in transfer-learning settings to reduce the number of trainable parameters and accelerate adaptation. Encoder-only or decoder-only fine-tuning was not retained, since the objective is to adapt both the latent manifold and the associated reconstruction mapping consistently to the new dynamical regime. Updating only the encoder would modify the latent topology while leaving the decoder incompatible, whereas fine-tuning only the decoder assumes that the latent representation remains adequate despite the distribution shift. We also evaluated a partial layer-freezing strategy in which only the deepest encoder layers and shallowest decoder layers were fine-tuned. Although this preserves encoder-decoder symmetry, it remained significantly less effective than full VAE fine-tuning. This limitation is attributed to the parametric conditioning being introduced at the encoder input, such that the early encoder layers play a key role in constructing the parameter-aware latent representation. Restricting adaptation to shallow latent-space layers therefore prevents the network from properly updating the feature--parameter fusion mechanism.
 
\subsection{Kalman Uncertainties}
\label{subsec:second_moment}

The Kalman filter produces an \textit{analysis ensemble} that inherently carries a certain degree of uncertainty. The covariance of this ensemble provides a quantitative measure of that uncertainty. Therefore, an important question is whether the model’s uncertainty estimate should explicitly incorporate the uncertainty already present in the Kalman filter’s analysis ensemble. In other words, should the fine-tuning process account not only for the first statistical moment (the mean state) but also for the second moment (the covariance) of the assimilated data? In our current framework, we enforce agreement between the model’s mean prediction and the analysis-ensemble mean, thereby aligning the first moments. The remaining issue is whether we should also align the second moments by producing stochastic predictions whose covariance captures both the model’s intrinsic uncertainty and the \textit{aleatoric} uncertainty introduced by the Kalman filter.

The following section demonstrates that such an additional treatment of the second moment is not required in our framework. Note, however, that other works (e.g. \citep{bocquet_2}) include the second moment in their retraining objective by penalizing or rewarding more confident states through their loss formulation (Equation~\ref{eq:alternative_loss}). Unlike our stochastic framework, their strategy is deterministic, yet this highlights that properly accounting for Kalman filter uncertainties remains crucial for building a trustworthy Machine Learning–DA hybrid system.

\paragraph{Encoder Definition}
We define the encoder operator $\mathcal{E}$, which maps a high-dimensional physical state to a probabilistic latent representation parameterized by its mean and (diagonal) covariance:
\begin{equation}
    \mathcal{E} : \mathbb{R}^{m} \rightarrow \mathbb{R}^{2 \times p}, 
    \qquad 
    \mathcal{E}(\psi) = \big( \mu_{\theta}, \mathrm{diag}(\Lambda_{\theta}) \big),
    \qquad
    \text{with } p \ll m.
\end{equation}
The subscript $\theta$ denotes the network parameters. The vector $\mathrm{diag}(\Lambda_{\theta})$ represents the encoder-predicted variance (the second statistical moment). Conceptually, this variance acts as a proxy for the diagonal entries of the true posterior covariance matrix of $p(z \,|\, \psi)$, where $z$ denotes the latent coordinates. The VAE therefore assumes statistical independence across latent dimensions, neglecting off-diagonal covariance components.

\paragraph{First-Moment Matching}
Fine-tuning enforces alignment between the first moments: the mean forecast of the model is matched to the mean of the analysis ensemble, denoted by $X^a$. This matching is achieved via the loss function in Equation~\ref{eq:loss}. The next step is to verify that the model remains insensitive to the covariance structure of the analysis ensemble.

\paragraph{Definition of the Analysis Ensemble}
Let the updated (assimilated) forecast ensemble be
\[
\psi^a = \{\psi_1^a, \psi_2^a, \dots, \psi_N^a\} \in \mathbb{R}^{N \times m},
\]
where each $\psi_i^a$ is a sample drawn from the posterior distribution
\[
\psi_i^a \sim \mathcal{N}(X^a, P^a),
\]
and where
\[
X^a = \frac{1}{N} \sum_{i=1}^N \psi_i^a, 
\qquad 
P^a = \mathrm{Cov}(\psi^a)
\]
denote the ensemble mean and its empirical covariance, respectively.

\paragraph{Latent-Space Representation}
Each ensemble member $\psi_i^a$ is mapped into the latent space via the encoder:
\begin{equation}
    \mathcal{E}: \psi_i^a \mapsto z_i^a, \quad z_i^a \in \mathbb{R}^p,
\end{equation}
yielding a latent ensemble $
    z^a \sim \mathcal{N}(\mu_z^a, \Sigma_z^a),
$
with empirical mean and covariance
\[
\mu_z^a = \frac{1}{N} \sum_{i=1}^N z_i^a, 
\qquad 
\Sigma_z^a = \frac{1}{N-1} \sum_{i=1}^N (z_i^a - \mu_z^a)(z_i^a - \mu_z^a)^\top.
\]
This ensemble thus defines fixed first and second moments in the latent space, providing a reference distribution for the autoencoder to reproduce.

\paragraph{Variational Encoding}
Within the VAE framework, the encoder outputs a Gaussian approximation to the posterior:
\begin{equation}
    \mathcal{E} : X^a \mapsto 
    \hat{z}_\theta^a \sim 
    \mathcal{N}\!\left(\mu_\theta^a,  \mathrm{diag}(\Lambda_\theta^a\right)),
\end{equation}
where $\mu_\theta^a \in \mathbb{R}^p$ is the predicted latent mean, and 
$\Lambda_\theta^a \in \mathbb{R}^{p \times p}$ is the predicted (diagonal) latent covariance.

Since the latent means are already aligned (\(\mu_\theta^a = \mu_z^a\); see Section~\ref{subsec:DA_framework}), our focus now shifts to ensuring that the predicted covariance \(\Lambda_\theta^a\) approximates the latent covariance structure \(\Sigma_z^a\) obtained from the assimilated ensemble. We thus compare:
\[
\Sigma_z^a =
\begin{pmatrix}
\sigma_1^2 & \sigma_1\sigma_2 & \cdots & \sigma_1\sigma_p \\
\sigma_2\sigma_1 & \sigma_2^2 & \cdots & \sigma_2\sigma_p \\
\vdots & \vdots & \ddots & \vdots \\
\sigma_p\sigma_1 & \sigma_p\sigma_2 & \cdots & \sigma_p^2
\end{pmatrix},
\qquad
\Lambda_\theta^a =
\begin{pmatrix}
\lambda_1 & 0 & \cdots & 0 \\
0 & \lambda_2 & \cdots & 0 \\
\vdots & \vdots & \ddots & \vdots \\
0 & 0 & \cdots & \lambda_p
\end{pmatrix},
\]
where each $\lambda_i$ is directly predicted by the encoder.

\paragraph{Minimization Problem}
If the VAE is designed to fit both the first and second statistical moments of the assimilated latent distribution, it must minimize the following constrained optimization problem:
\begin{equation}
    \label{eq:optimization_problem}
    \Lambda_\theta^{a*} 
    = \arg \min_{\Lambda_\theta^a} 
    D_{\mathrm{KL}}\!\left(
    \mathcal{N}(\mu_z^a, \Sigma_z^a)
    \,\|\, 
    \mathcal{N}(\mu_\theta^a, \Lambda_\theta^a)
    \right)
    \quad \text{subject to} \quad 
    (\Lambda_\theta^a)_{ij} = 0 \;\; \text{for } i \neq j.
\end{equation}
The Kullback–Leibler divergence between two multivariate normal distributions is:
\begin{equation}
\begin{aligned}
D_{\mathrm{KL}}\big(
\mathcal{N}(\mu_z^a, \Sigma_z^a)
\,\|\, 
\mathcal{N}(\mu_\theta^a, \Lambda_\theta^a)
\big)
= \tfrac{1}{2} \Big[
& \log \tfrac{\det \Lambda_\theta^a}{\det \Sigma_z^a} - p 
+ \mathrm{tr}\!\big((\Lambda_\theta^a)^{-1}\Sigma_z^a\big)
+ (\mu_\theta^a - \mu_z^a)^\top (\Lambda_\theta^a)^{-1} (\mu_\theta^a - \mu_z^a)
\Big].
\end{aligned}
\end{equation}
Since the first moments are already matched (\(\mu_\theta^a = \mu_z^a\)), this simplifies to
\begin{equation}
D_{\mathrm{KL}} =
\tfrac{1}{2} \left[
\log \tfrac{\det \Lambda_\theta^a}{\det \Sigma_z^a} - p 
+ \mathrm{tr}\!\big((\Lambda_\theta^a)^{-1}\Sigma_z^a\big)
\right].
\end{equation}
 Since $\Lambda_\theta$ is diagonal, we obtain:
\begin{equation}
D_{\mathrm{KL}} = 
\tfrac{1}{2} \left[
\log \tfrac{\prod_{i=1}^p \lambda_i}{\det \Sigma_z^a} 
- p + 
\sum_{i=1}^p \tfrac{\sigma_i^2}{\lambda_i}
\right].
\end{equation}
Ignoring constants, the minimization reduces to:
\[
\min_{\lambda_i > 0} \; f(\lambda_i) 
= \log\!\left(\prod_{i=1}^p \lambda_i\right) 
+ \sum_{i=1}^p \frac{\sigma_i^2}{\lambda_i}.
\]
Setting the derivative to zero gives:
\[
\frac{\partial f}{\partial \lambda_i} 
= \frac{1}{\lambda_i} - \frac{\sigma_i^2}{\lambda_i^2} = 0
\quad \Rightarrow \quad 
\lambda_i^* = \sigma_i^2.
\]
\[
\left. \frac{\partial^2 f}{\partial \lambda_i^2} \right|_{\lambda_i = \lambda_i^*} = -\frac{1}{\lambda_i^{2^*}} + \frac{2\sigma_i^2}{\lambda_i^{3^*}} = \frac{2\sigma_i^2 - \lambda_i^*}{\lambda_i^{3^*}}
\] 
Since \(\sigma_i^2 > 0\) and \(\lambda_i^* = \sigma_i^2\), we have
\[
\left. \frac{\partial^2 f}{\partial \lambda_i^2} \right|_{\lambda_i = \lambda_i^*} > 0,
\]
which shows that the function \(f\) is locally convex around \(\lambda^*\), making \(\lambda^*\) a local minimum. Moreover, considering the monotonicity of the \(\log\) and inverse functions together with the behaviour of \(f(\lambda)\) tending to infinity as \(\lambda_i \to 0^+\) or \(\lambda_i \to +\infty\), we conclude that \(f\) admits a unique global minimum. 
Thus, the optimal diagonal covariance approximation is:
\[
\Lambda_\theta^{a*} = \mathrm{diag}(\Sigma_z^a).
\]

\begin{figure}[H]
    \centering
    \includegraphics[width=0.8\textwidth]{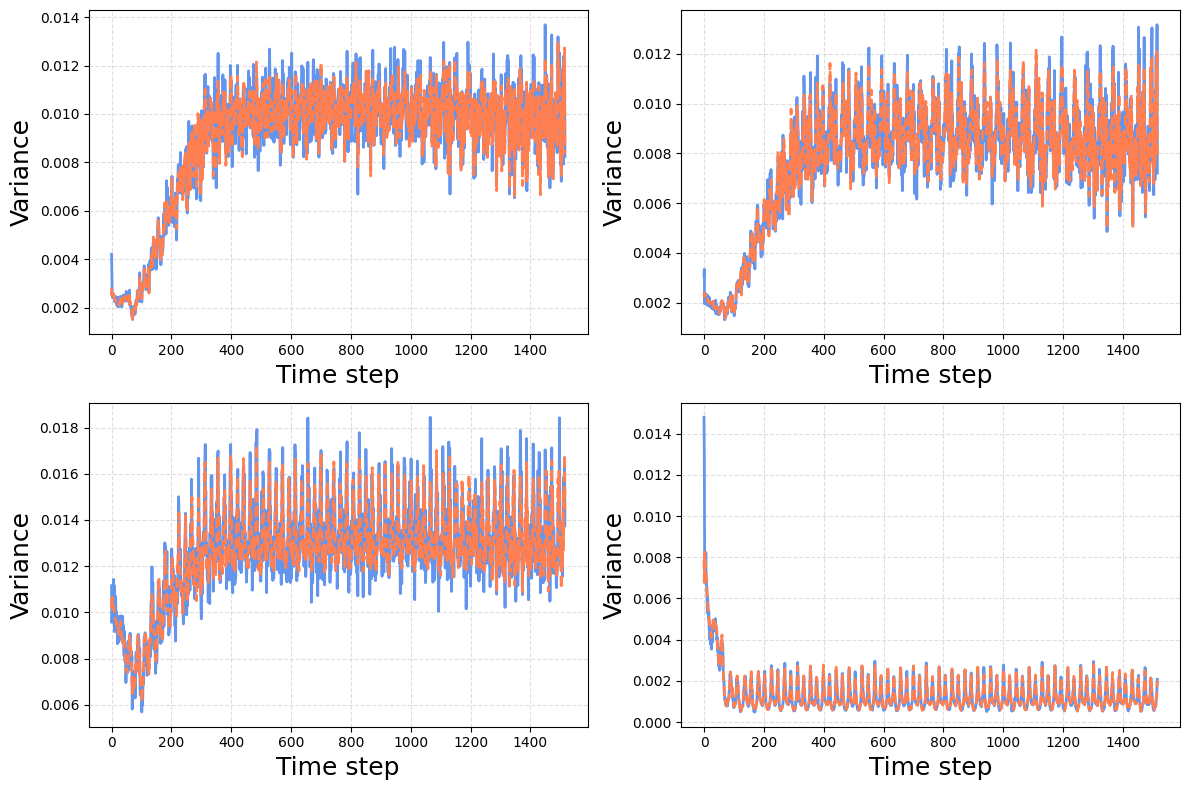}
    \includegraphics[width=0.4\textwidth]{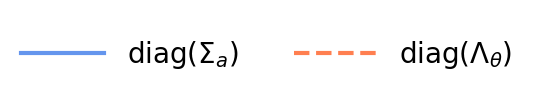}
    \caption{Comparison of $p$ pairs of variance signals: true latent variances (from encoding the ensemble) versus variance predictions from the encoder (based on the ensemble mean).}
    \label{fig:variancevssigma_VAE}
\end{figure}

\paragraph{Interpretation}
This result implies that if encoding the mean of the analysis ensemble, $X^a$, yields variance components $\lambda_{\text{VAE}} =  \mathrm{diag}(\Lambda_\theta^a)$ that, at each time step, are equal to the diagonal entries of the covariance matrix obtained by encoding each ensemble member individually, then the model’s second statistical moment faithfully captures that of the Kalman filter. Formally,
\begin{equation}
\label{eq:optimization_pb}
    \lambda_{\text{VAE}}^* 
    = \arg\min_{\lambda_{\text{VAE}}} 
    \left\| 
        \lambda_{\text{VAE}} - 
        \mathrm{diag}\!\left(\mathrm{Cov}\!\left[\mathcal{E}(\psi_i^a)\right]\right)
    \right\|.
\end{equation}
In our experimental setup with $p=4$, we can confirm that the optimization problem defined in Equation~\ref{eq:optimization_pb} does not need to be solved through tailored retraining as, after the initial, first moment fine-tuning, we already have consistent agreement of the second moment statistics. Figure~\ref{fig:variancevssigma_VAE} presents a comparison of the $p=4$ signals, plotted two by two, between the true latent variances obtained from the ensemble and the corresponding variance estimates produced by the encoder when applied to the ensemble mean.  

These results demonstrate that the VAE effectively captures the uncertainty structure of the analysis ensemble without requiring the explicit inclusion of second-moment terms in the loss function. This statement holds true within our experimental setup, but it is not necessarily guaranteed to generalize to other systems. However, as a rule of thumb, when the analysis ensemble variance remains small compared with the system’s total energy, such an approximation can be reasonably made. 
Since we only consider the diagonal components of the analysis ensemble covariance matrix, that is, its variance, this uncertainty can be interpreted as a form of additive noise. Consequently, the model can effectively suppress this noise through its inherent denoising capability. 

Finally it should be noted that even though the noise does not affect  the methodology, it could impact fine-tuning performances, particularly for the second moment. To address this, smoothing filters can be applied prior to fine-tuning to effectively denoise the assimilated data, potentially improving the performance of the DA–DL framework. 
However, designing such filters remains challenging for high-dimensional, multi-scale dynamics and nonlinear systems, particularly when the predictive model operates as a black box that is incompatible with classical smoothers such as the Rauch–Tung–Striebel smoother.

\section{Conclusion}
\label{sec:conclu}

This work introduces an efficient fine-tuning methodology  for parametric dynamical models. The proposed approach is intended for situations in which the model must adapt to a new dynamical regime that remains sufficiently close to the training distribution. In particular, the method assumes that the underlying dynamics preserve the same qualitative behaviour between the training and adaptation datasets. Therefore, the transition should not involve major dynamical changes such as bifurcations, radical attractor modifications, quasiperiodic behaviour, intermittency, or transitions to chaos. Under these assumptions, namely within a fixed dynamical regime, the adaptation process can be significantly simplified and accelerated. Efficiency is achieved through two key mechanisms: (i) fine-tuning only a designated subcomponent of the model, which substantially reduces training time, and (ii) incorporating sparse datasets during the fine-tuning process. 

Firstly, by only adapting the components of the model that require correction, namely, the parts of the learned manifold that shift between nearby Reynolds regimes, we can avoid making unnecessary updates to the underlying dynamics, which remain largely unchanged. Empirically, when flow conditions differ only moderately, the manifold deforms while the trajectory on it remains preserved, enabling rapid and inexpensive fine-tuning. Secondly, since acquiring data is costly, we demonstrate that accurate adaptation can be achieved using a very small amount of data via assimilation enabled by ensemble Kalman filters. These leverage the stochastic nature of our reduced order model, resulting in a symbiotic relationship between data assimilation and machine learning. The model's stochastic and Gaussian properties benefit the assimilation process, while assimilation generates new data to retrain and improve the model.

This selective adaptation results in near-real-time convergence for both the first and second moments, as well as updates within seconds when only the first moment is required. Ultimately, by using just one percent of the full-state data, acquired strategically via targeted sensor placement, we can reduce prediction error by up to 70\%, representing an excellent trade-off between performance and fine-tuning cost. 
While data assimilation enables this efficient correction, it also introduces noise that can reduce the model’s confidence. This decreased certainty arises from the assimilation noise itself rather than from the intrinsic EnKF uncertainty, and therefore does not necessitate incorporating ensemble variance into the fine-tuning step. In principle, smoothing filters could be used to mitigate this noise and improve robustness further.

Overall, the proposed framework strikes a favourable balance between accuracy, computational cost, and data requirements, providing a practical and scalable avenue for fast domain adaptation in learned dynamical models.

\bibliography{references.bib}

\paragraph{Funding Statement}
This project is co-funded by the European Union’s Horizon Europe research and innovation programme Cofund SOUND.AI under the Marie Sklodowska-Curie Grant Agreement No 101081674. 

\paragraph{Competing Interests}
None

\paragraph{Data Availability Statement}
The Reduced Order Model (ROM) is openly available on GitHub \cite{GitHubRepo} at \url{https://github.com/IsmaelZig-SU/p-DynSys_EncoderTransformerDecoder}
. The complete modelling pipeline, including the source code and experimental datasets used in this study, is available in a separate GitHub repository \cite{OuroborosModel} at \url{https://github.com/IsmaelZig-SU/OuroborosModel}.

\paragraph{Ethical Standards}
The research meets all ethical guidelines, including adherence to the legal requirements of the study country.

\begin{appendix}
\section{Appendix}
\label{sec:Appendix}

\subsection{ROM architecture}
\label{subsec:ROMarch}

\begin{table}[H]
\centering
\begin{tabular}{ll}
\toprule
\textbf{Parameter} & \textbf{Value} \\
\midrule
Data parametric dimensions (train) & 2 \\
Data parametric dimensions (validation/test) & 7 \\
Time dimension & 1516 \\
Spatial dimensions & $131 \times 100 \div 4 = 3275$ \\
Flows & $u, v$ \\
Input dimension & $3275 \times 2 = 6550$ \\
Latent dimensions & 4 \\
Transformer hidden dimensions & 64 \\
Kullback-Leibler regularization & $3 \times 10^{-4}$ \\
Attention blocks & 1 \\
Attention heads & 8 \\
Prediction horizon & 10 \\
Lookback window & 9 \\
\bottomrule
\end{tabular}
\caption{Model and data configuration parameters.}
\end{table}

The current study uses 4 latent dimensions. From a purely dynamical systems perspective, a two-dimensional latent space is theoretically sufficient to represent a single periodic orbit. However, the present framework must also capture transient dynamics toward the attractor, Reynolds-number variations, and the stochastic nature of the VAE latent representation. In practice, the latent space is not optimized solely for reconstruction accuracy, but also for the probabilistic consistency imposed by the variational formulation. These additional constraints require a richer latent representation.

\begin{figure}[H]
    \centering
    \includegraphics[width=0.5\linewidth]{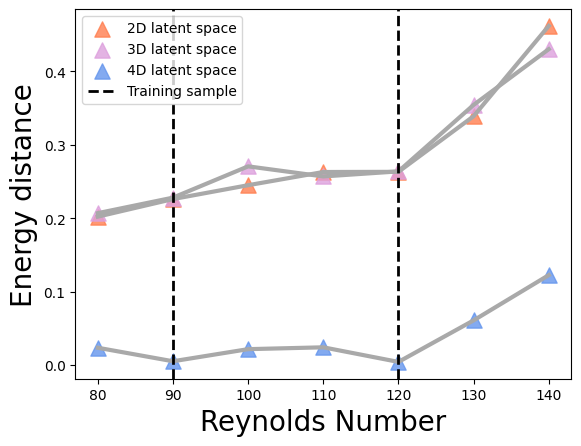}
    \caption{Energy distance across the Reynolds-number range for latent dimensions $p=2$, $p=3$, and $p=4$.}
    \label{fig:ablation}
\end{figure}

To quantitatively justify this choice, we performed a short ablation study comparing latent dimensions $p=2$, $p=3$, and $p=4$. Figure~\ref{fig:ablation} reports the energy distance across the Reynolds-number range. A significant reduction of the energy distance is observed for $p=4$. This suggests that four latent dimensions provide the minimum latent-space complexity required to accurately capture both the dynamics and uncertainty structure of the system.

Furthermore, using four latent dimensions may facilitate future extensions toward more complex dynamical regimes involving additional periodic orbits at higher Reynolds numbers. This will be the object of future studies.

The selection of the look-back window ($q$) and prediction horizon ($H$) is guided by two main constraints. First, to maximize computational efficiency, both $q$ and $H$ are minimized to reduce the memory footprint and accelerate training and inference while maintaining high predictive accuracy. Second, to ensure rigorous autoregressive evaluation we enforce $m > q$. This setup forces the model to predict at least one future state using a context window composed entirely of its own historically generated predictions, ensuring robustness during rollouts through auto-regressive feedback-loop.

\subsection{Sensor location}
\label{subsec:sensors}
Conventional methods in data compression, sparse reconstruction, and sparse optimization involve estimating the high-dimensional state using low-rank approximations. For instance, when dealing with flow past an obstacle, the data can often be accurately represented using a few eigenmodes through Singular Value Decomposition (SVD). By employing tailored sensing, we can optimize the placement of observations using a subsampling matrix that captures as much information as possible about the full state signal in the most sparse manner, making it a classic optimization problem~\cite{sensor_placement}. 

The low rank approximation of a full-state, high dimensional, potentially noisy vector $x \in \mathbb{R}^{T \times m}$ with $T$ and $m$ the time dimension and spatial dimension respectively is derived below.

Starting from the Singular Value Decomposition (SVD), we write : 
\begin{equation}
x = U \Sigma V^T,
\end{equation}
where \( U \in \mathbb{R}^{m \times k} \) contains the spatial modes, \( \Sigma \in \mathbb{R}^{k \times k} \) is the diagonal matrix of singular values, and \( V \in \mathbb{R}^{T \times k} \) holds the temporal coefficients, with \( k \leq \min(T, D) \). We define \( \Psi = U \Sigma \), which combines the spatial structures with their energetic weights.
\begin{equation}
\label{eq:svd}
    x \approx x_r = \Psi_r a
\end{equation}
where the subscript $r$ indicates the rank of truncation in the eigenbase, $a$ indicates the "mixture" of active modes. We also define the sub-sampling operator $H \in \mathbb{R}^{p \times n}$, which defines the locations of the $p$ sensors across the full state data $x \in \mathbb{R}^n$. The observations $y$ are therefore defined as:
\begin{equation}
\label{eq:subsamp}
    y = H x
\end{equation}
Combining equations \ref{eq:svd} and \ref{eq:subsamp} implies that:
\begin{equation}
    y = H \Psi_r a
\end{equation}
This means that we could theoretically reconstruct the full state data by simply computing $a$ from the sensors observations via:
\begin{equation}
a = 
\begin{cases}
    (H \Psi_r)^{-1} y & \text{if } n = r \\
    (H \Psi_r)^\dagger y & \text{if } n > r
\end{cases}
\end{equation}
In practice, the reconstruction of \( a \) heavily depends on the structure of \( H \). Specifically, it is crucial to ensure that minor noise in \( y \) does not drastically affect the signal reconstruction. Ensuring that small variations in \( y \) result in small variations in the full state reconstruction is a problem known as condition number minimization. This means that the inversion of \( H \Psi_r \) should be well-conditioned and not rank-deficient. \\
Equivalently, we seek the structure of $H$ that maximizes the determinant of $H \Psi_r$ 
\begin{equation}
   H = \arg\max_H \left| \det(H \Psi_r) \right|
\end{equation}
leading to,
\begin{equation}
   H = \arg\max_H \prod_{i=1}^r \lambda_i(H \Psi_r)
\end{equation}
where $\lambda_i$ are the eigenvalues of the matrix product $H \Psi_r$. This remains a combinatorial NP-hard problem, but a greedy approximation of $H$ can be found from the pivot matrix $C$ in the QR factorization of the basis matrix $\Psi$ \cite{sensor_placement}
\begin{equation}
    \Psi_r^T C^T = Q R
\end{equation}
where \( C \) is the pivot matrix containing \( p = r \) non-zero entries, \( Q \) is an orthogonal, unitary matrix and \( R \) is an upper triangular matrix. This latter statement holds true as long as \( p = r \), but other formulations can be derived in the case where \( p > r \). The pivot matrix \( C \) iteratively finds the rows of \( \Psi_r \) with the highest \( l_2 \)-norm, the orthogonal projection onto this pivot row is then subtracted from the other rows before repeating the procedure. We can then have a sense of why \( C \) is the best greedy approximation for \( H \) as it finds the placements with the highest modal variance, thus carrying the most information about each mode. This iterative process ensures that each diagonal entry in \( R \) is as large as possible given the preceding entry. Since \( R \) is a triangular matrix, we have:
\begin{equation}
    \det(\Psi_r^T C_r^T) = \det(\Psi_r C_r) = \prod_{i=1}^{r=p} R_{ii} \approx \max_H \left| \det(H \Psi_r) \right|.
\end{equation}

To place the sensors, we perform a QR-pivoted factorization on the ensemble forecast mean field $\mathbf{X}^f$ derived from the forecast ensemble $\psi^f$. Since the decomposition is applied exclusively to the predicted state and not to the ground-truth target solution, the sensor placement procedure remains entirely model-driven. Consequently, the selected sensor locations are determined solely from the forecast information, ensuring consistency with the sparse-observation framework considered in this work.
To motivate the choice of QR factorization over other placement algorithms, we conducted a comparative analysis. 
We considered four distinct sensor placement strategies to reconstruct the flow field defined by the velocity components $U$ and $V$ with 62 sensors :

\begin{enumerate}
    \item QR Factorization (QR-pivoting) :
    
    \begin{figure}[H]
    \centering
    \begin{subfigure}[t]{0.48\textwidth}
        \includegraphics[width=1\linewidth]{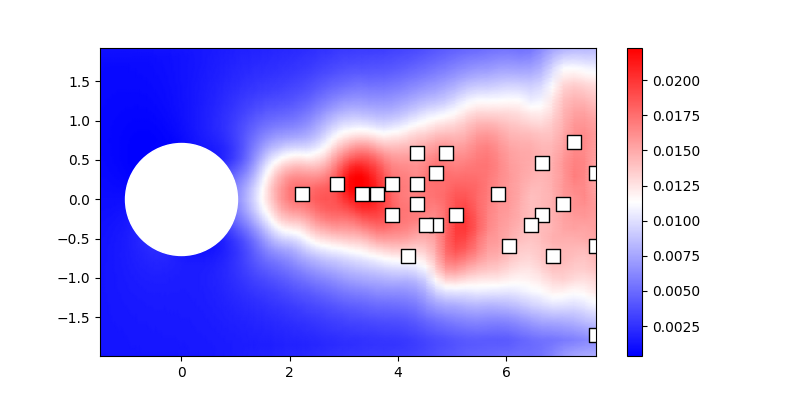}
        \caption{Sensor locations for $U$ velocity field, represented on the average U uncertainty with QR factorization}
        \label{fig:sensors_U_a}
    \end{subfigure}
    \begin{subfigure}[t]{0.48\textwidth}
        \includegraphics[width=1\linewidth]{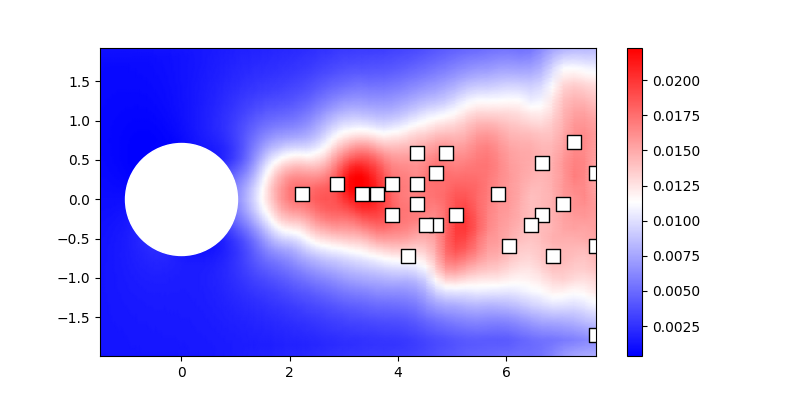}
        \caption{Sensor locations for $V$ velocity field, represented on the average V uncertainty with QR factorization}
        \label{fig:sensors_V_a}
    \end{subfigure}
    \caption{Sensor placement computed with QR factorization.}
    \label{fig:sensors_a}
    \end{figure} 
    
    \item Maximum Variance (Max-Var): \\
    Sensors are placed at the locations exhibiting the highest predictive uncertainty. Let $\sigma_i^2$ denote the variance of the predicted state at spatial location $i$, extracted from the forecast covariance matrix $P^f$. The selected sensor set $\mathcal{S}$ is obtained from
    \[
        \mathcal{S} = \arg\max_{|\mathcal{S}|=m} \sum_{i \in \mathcal{S}} \sigma_i^2 ,
    \]
    where $m$ is the number of sensors. This strategy prioritizes regions where the model uncertainty is largest, however it often results in sets where inner correlations are important, thus providing redundant information to the filter.

    \begin{figure}[H]
    \centering
    \begin{subfigure}[t]{0.48\textwidth}
        \includegraphics[width=1\linewidth]{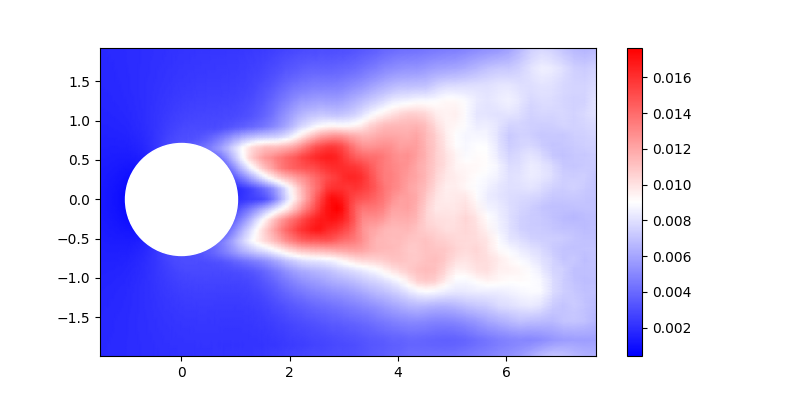}
        \caption{Sensor locations for $U$ velocity field, represented on the average U uncertainty with maximum variance}
        \label{fig:sensors_U_b}
    \end{subfigure}
    \begin{subfigure}[t]{0.48\textwidth}
        \includegraphics[width=1\linewidth]{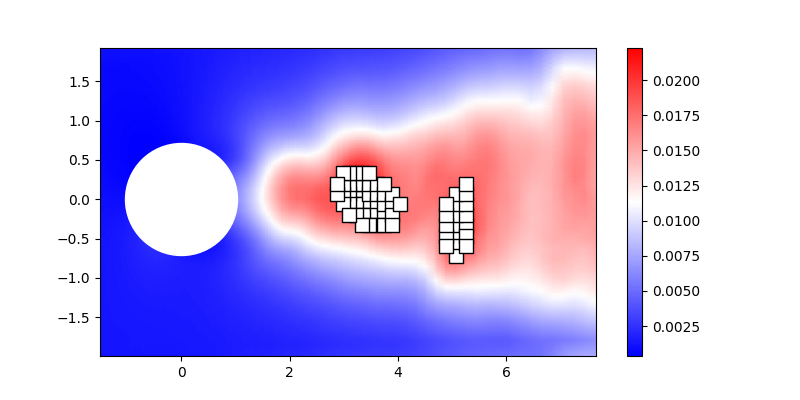}
        \caption{Sensor locations for $V$ velocity field, represented on the average V uncertainty with maximum variance}
        \label{fig:sensors_V_b}
    \end{subfigure}
    \caption{Sensor placement computed with maximum variance.}
    \label{fig:sensors_b}
    \end{figure} 
    
    \item Maximum Variance -- Minimum Inner-Set Covariance (Max-Var Min-Corr) : \\
    This strategy extends the previous approach by seeking sensor locations that both maximize uncertainty and minimize redundant information. A greedy optimization procedure is employed to construct the sensor subset iteratively. Denoting by $P^f_{ij}$ the covariance between locations $i$ and $j$, the optimization problem can be formulated as
    \[
        \mathcal{S}^\star
        =
        \arg\max_{|\mathcal{S}|=m}
        \left(
        \sum_{i\in\mathcal{S}} \sigma_i^2
        -
        \lambda
        \sum_{\substack{i,j\in\mathcal{S}\\ i\neq j}}
        P^f_{ij}
        \right),
    \]
    where $\lambda$ controls the trade-off between uncertainty maximization and redundancy penalization (here $\lambda = 0.2)$. The resulting sensor network captures highly informative yet weakly correlated measurements.

    \begin{figure}[H]
    \centering
    \begin{subfigure}[t]{0.48\textwidth}
        \includegraphics[width=1\linewidth]{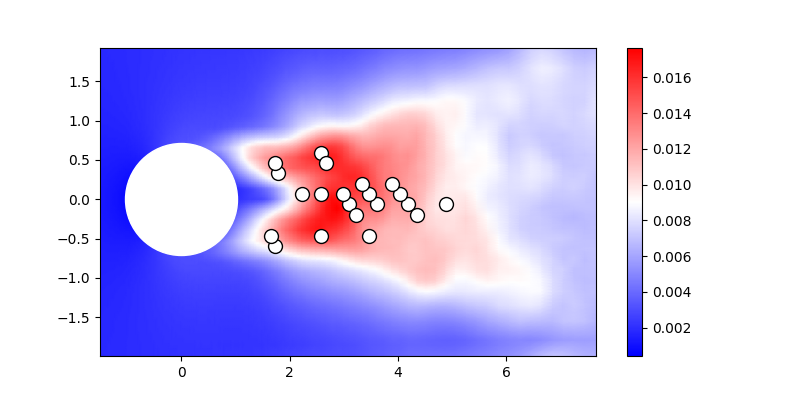}
        \caption{Sensor locations for $U$ velocity field, represented on the average U uncertainty with maximum variance - minimum inner-set covariance}
        \label{fig:sensors_U_c}
    \end{subfigure}
    \begin{subfigure}[t]{0.48\textwidth}
        \includegraphics[width=1\linewidth]{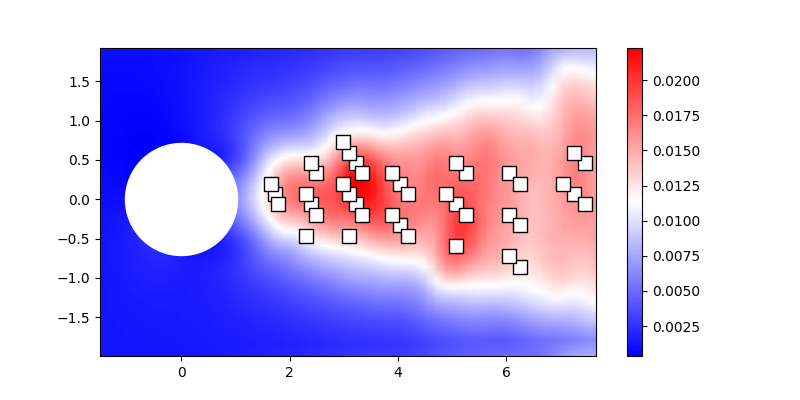}
        \caption{Sensor locations for $V$ velocity field, represented on the average V uncertainty with maximum variance - minimum inner-set covariance}
        \label{fig:sensors_V_c}
    \end{subfigure}
    \caption{Sensor placement computed with maximum outer variance - minimum inner-covariance.}
    \label{fig:sensors_c}
    \end{figure}

    \item Random Placement : \\ 
    Sensor locations are sampled uniformly at random over the spatial domain. Three independent realizations are considered to provide a statistical baseline for comparison against the deterministic placement strategies.

        \begin{figure}[H]
        \centering
        \begin{subfigure}[t]{0.48\textwidth}
            \includegraphics[width=1\linewidth]{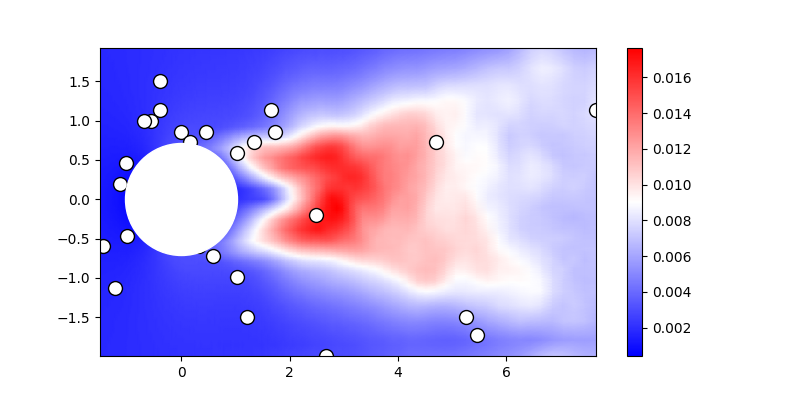}
            \caption{Sensor locations for $U$ velocity field, represented on the average U uncertainty with maximum variance - minimum inner-covariance}
            \label{fig:sensors_U_d}
        \end{subfigure}
        \begin{subfigure}[t]{0.48\textwidth}
            \includegraphics[width=1\linewidth]{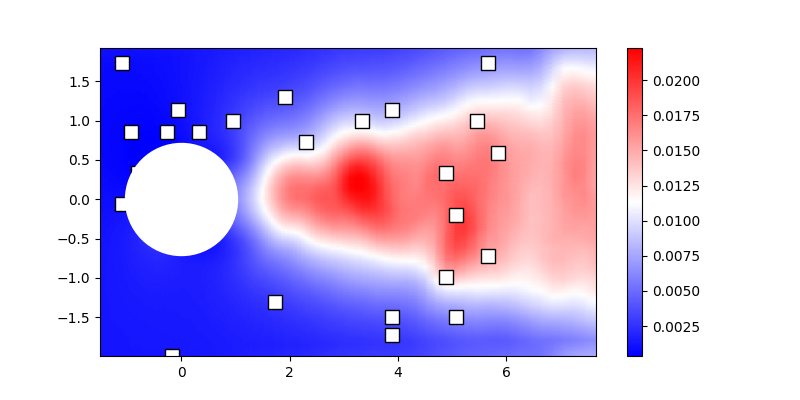}
            \caption{Sensor locations for $V$ velocity field, represented on the average V uncertainty with maximum outer variance - minimum inner-covariance}
            \label{fig:sensors_V_d}
        \end{subfigure}
        \caption{Sensor placement computed with maximum variance - minimum inner-covariance.}
        \label{fig:sensors_d}
        \end{figure} 
    
\end{enumerate}

\begin{figure}[H]
    \centering
    \includegraphics[width=0.6\linewidth]{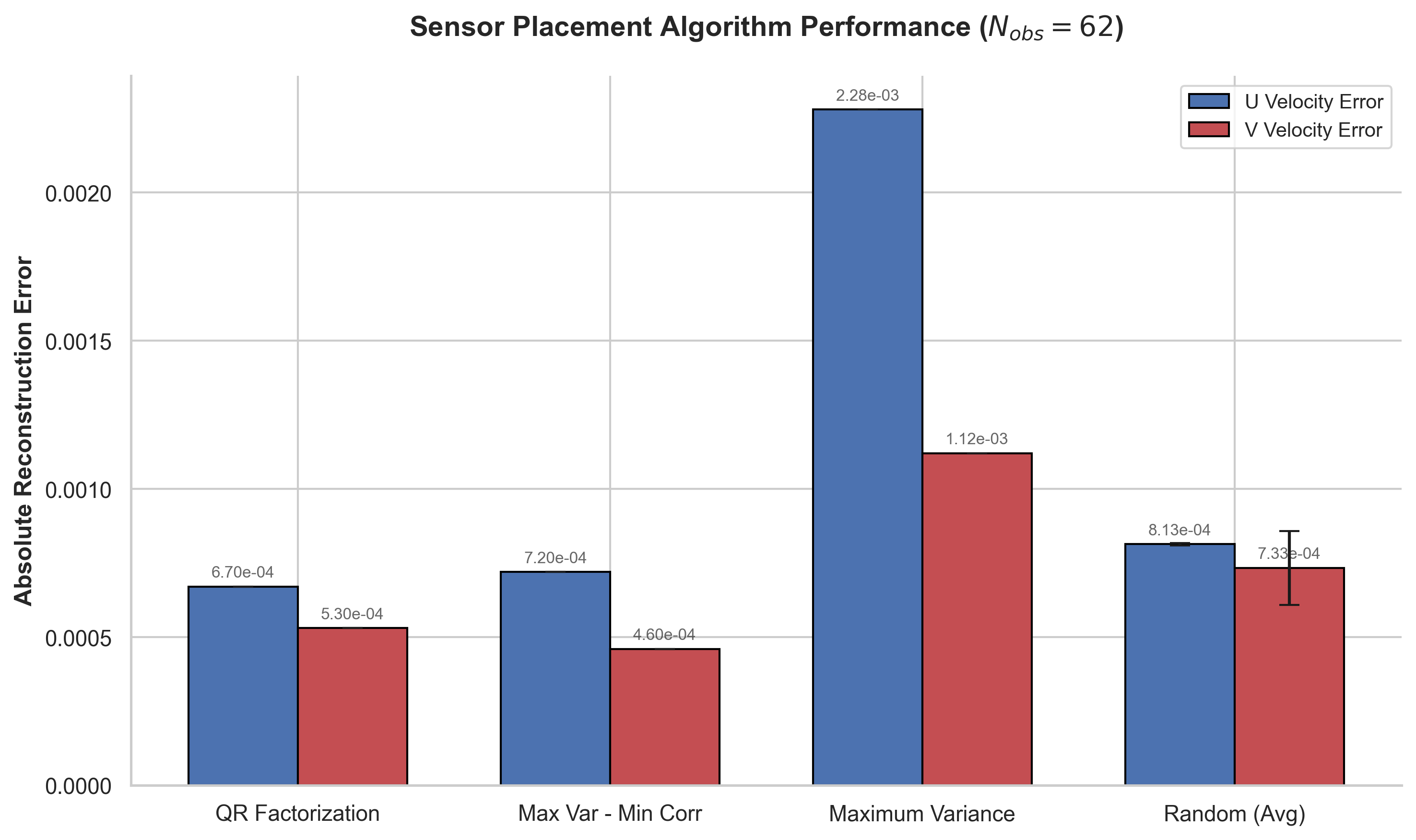}
    \caption{Comparison of the reconstruction error obtained after Kalman filtering using different sensor placement strategies. The random placement strategy was evaluated over three independent realizations in order to assess the variability and statistical robustness of random sensor configurations.}
    \label{fig:results_sensor_placement}
\end{figure}
The results, summarized in figure \ref{fig:results_sensor_placement} in the context of absolute reconstruction error, indicate that while the \textit{Maximum Variance} approach is prone to clustering and high error, the introduction of a decorrelation constraint in the \textit{Max Var - Min Corr} algorithm significantly improves performance. However, the \textit{QR Factorization} method yielded nearly identical performance to the optimized greedy approach. Given its computational efficiency and status as a standard for discrete sensor optimization, QR factorization was selected as the primary methodology for the data assimilation steps. 

To further assess the influence of the sensor configuration and ensemble cardinality on the assimilation performance, additional experiments were conducted using different numbers of sensors (\(S = 32, 64, 128\)) and a larger ensemble size (\(N = 100\)). The results, summarized in Table~\ref{tab:cardinality_vs_RNMSE}, report the global Relative Root Mean Squared Error (RRMSE) associated with the forecasted field \(\psi^f\) and the assimilated field \(\psi^a\).

The comparison shows that increasing the number of sensors generally improves the assimilation accuracy, although the gain becomes progressively smaller beyond \(S=64\). Similarly, enlarging the ensemble cardinality from \(N=20\) to \(N=100\) only provides moderate improvements in the assimilated RRMSE. These results suggest that the proposed framework already achieves stable and accurate corrections with relatively small ensembles and a moderate number of sensors, making \(N=20\) an effective compromise between computational cost and assimilation accuracy.

\begin{table}[H]
  \centering
  \caption{Global RRMSE comparison of forecast and assimilated ensembles for different ensemble cardinalities and sensor configurations.}
  \label{tab:cardinality_vs_RNMSE}
  \begin{tabular}{@{}rrcc@{}}
    \toprule
    $N$ & $S$ & RRMSE forecast [\%] & RRMSE assimilated [\%] \\
    \midrule
    20  & 32  & 8.15 & 2.56 \\
    20  & 64  & 8.15 & 2.31 \\
    20  & 128 & 8.15 & 2.31 \\
    100 & 32  & 7.97 & 2.42 \\
    100 & 64  & 7.97 & 1.97 \\
    100 & 128 & 7.97 & 1.87 \\
    \bottomrule
  \end{tabular}
\end{table}

\subsection{Gaussianity validation of the generated forecast ensemble}
\label{subsec:gaussianity}

\begin{figure}[H]
    \centering
    \includegraphics[width=0.6\linewidth]{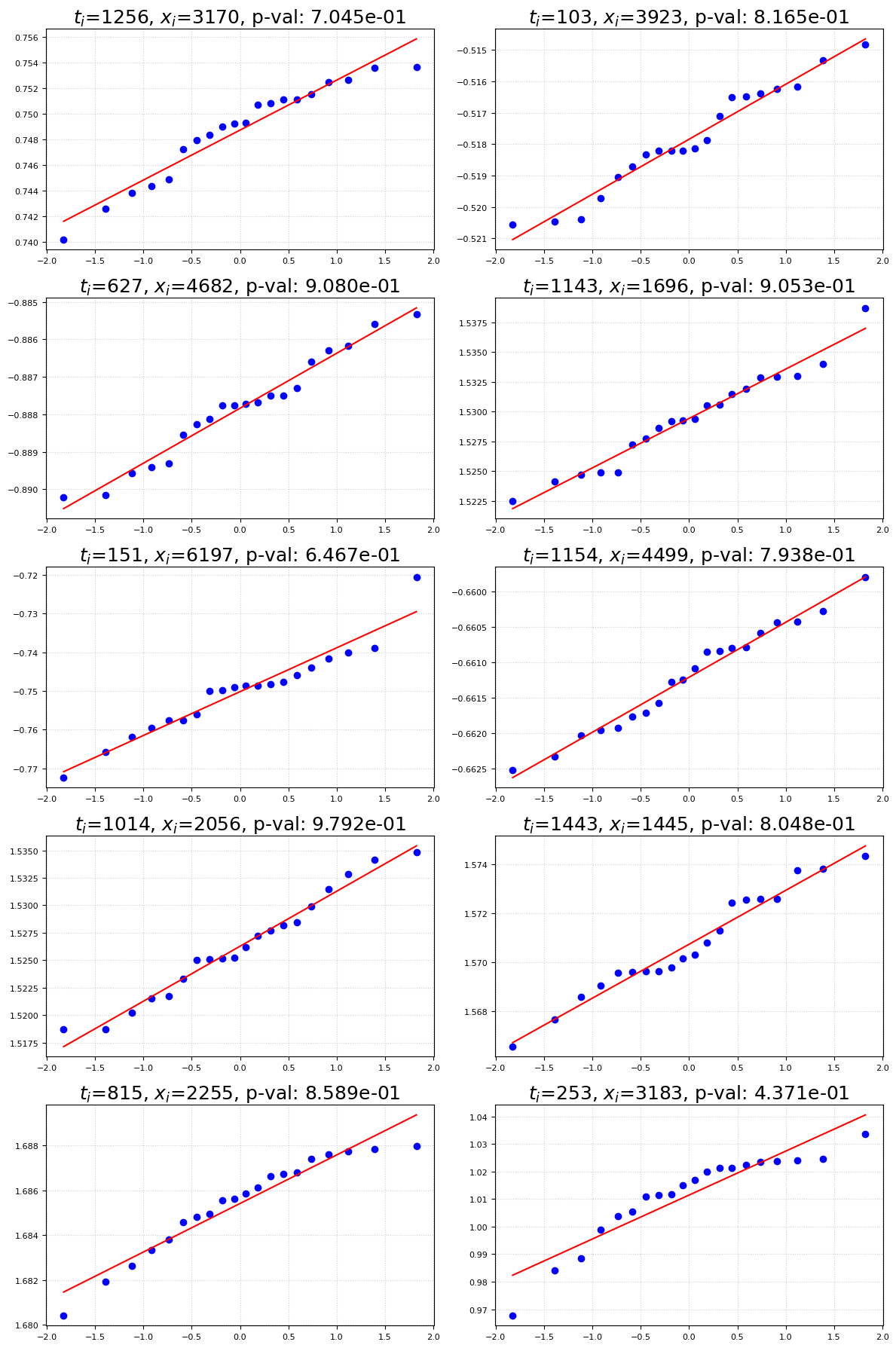}
    \caption{Q--Q plots for 20 randomly sampled spatial locations and temporal snapshots of the forecast ensemble $\psi^f_{t_i,x_j}$, together with the associated Gaussianity test $p$-values.}
    \label{fig:qqplot}
\end{figure}

To further assess the Gaussian assumption, we provide in Figure~\ref{fig:qqplot} a set of 10 Q-Q plots corresponding to randomly sampled snapshots $(t=t_i)$ and spatial locations $(x=x_j)$ of the forecast ensemble $\psi^f_{t_i,x_j}$. Each plot is accompanied by its associated $p$-value. In all tested cases, the obtained $p$-values remain above $5\times 10^{-3}$, while the empirical quantiles exhibit an overall satisfactory agreement with the theoretical Gaussian quantiles. Across the entire dataset, we concluded that the $p$-value exceeded that threshold is $99.04\%$ of the cases.

Although these results do not constitute a rigorous proof of Gaussianity, they provide a qualitative sanity check indicating that the ensemble distributions are sufficiently ``Gaussian-like'' for the application of the Kalman filter framework. In practice, the performance of Kalman filtering improves as the ensemble statistics approach Gaussianity. Consequently, localized departures from Gaussian behaviour at specific spatial points or time instants should primarily be viewed as potential limitations affecting optimality, rather than as conditions invalidating the filtering procedure altogether.

\end{appendix} 

\end{document}